\documentclass[11pt]{article}

\usepackage[final]{acl}

\usepackage{times}
\usepackage{latexsym}

\usepackage[T1]{fontenc}

\usepackage[utf8]{inputenc}

\usepackage{microtype}

\usepackage{inconsolata}

\usepackage{graphicx}
\usepackage{amsmath,amsfonts}
\usepackage{algorithmic}
\usepackage{graphicx}
\usepackage{textcomp}
\usepackage[table]{xcolor}
\usepackage{algorithm}
\usepackage{colortbl}
\usepackage{booktabs}
\usepackage{hyperref}
\usepackage[utf8]{inputenc}
\usepackage{lipsum}
\usepackage[most]{tcolorbox}
\tcbuselibrary{breakable}
\usepackage[most]{tcolorbox}

\usepackage{caption}
\usepackage{subcaption}
\usepackage{enumitem}

\usepackage{amssymb}
\usepackage{mathtools}
\usepackage{amsthm}

\definecolor{deepred}{RGB}{139,0,0}
\definecolor{lightred}{RGB}{255,230,230}
\usepackage{multirow}

\newcommand{\Examplebox}[2]{
  \begin{tcolorbox}[title=\textbf{#1}, 
  enhanced,
  breakable,
  colback=gray!5!white,
  colframe=gray!80!black,
  coltitle=white,
  fonttitle=\bfseries,
  boxrule=0.8pt,
  arc=6pt,
  left=6pt,
  right=6pt,
  top=6pt,
  bottom=6pt,
  colupper=black,
  sharp corners=south,
  borderline={0.8pt}{0pt}{gray!80!black},
  before upper={\parindent0em}]
    #2
  \end{tcolorbox}
}
\title{DCFA: Dual-view Causal-inspired Attribution for Failure Reasoning in LLM-based Multi-agent Systems}

\author{
 \textbf{Zehao Wang}\textsuperscript{1,2},
 \textbf{Lanjun Wang}\textsuperscript{2}\thanks{~~Corresponding Author.},
 \textbf{Shilong Jin}\textsuperscript{1,2},
 \textbf{Junjie Chen}\textsuperscript{1},
 \textbf{Yanghua Xiao}\textsuperscript{3,4}
\\
\\
 \textsuperscript{1}College of Intelligence and Computing, Tianjin University, Tianjin, China \\
 \textsuperscript{2}School of New Media and Communication, Tianjin University, Tianjin, China \\
 \textsuperscript{3}College of Computer Science and Artificial Intelligence, Fudan University, Shanghai, China \\
 \textsuperscript{4} Shanghai Key Laboratory of Data Science, Shanghai, China
\\
 {
   \texttt{\{wzhrslh, jinshilong\_2025\}@gmail.com}
 } \\
 {
   \texttt{\{wanglanjun, junjiechen\}@tju.edu.cn}, \texttt{shawyh@fudan.edu.cn}
 }
}

\begin{document}
\maketitle
\begin{abstract}
Large language model (LLM)-based multi-agent systems have experienced rapid growth in recent years. Despite their promise, such systems remain fragile, frequently exhibiting reasoning and coordination errors that can lead to system-level failures. 
Failure attribution in such systems relies on tracing natural language interactions among agents to identify the decisive error, which refers to the earliest action whose correction can reverse system failure. 
There are two key challenges: 1) Shallow attribution: Existing methods often capture only minor deviations, such as incomplete retrievals or formatting errors, which verification mechanisms can correct, while missing the decisive cause of system failure. 2) Contextual degradation: As the length of the system traces increases, the model’s reasoning ability rapidly deteriorates. 
To address these challenges, we propose DCFA, a training-free framework for failure attribution. DCFA integrates a global module that constructs structured causal-inspired dependency graphs from system traces to identify the initial decisive error, and a local module that applies local counterfactual-inspired reasoning to refine causal-inspired attribution.
Experiments on the Who\&When benchmark across six LLMs show that DCFA improves step-level accuracy by up to 8.27\% over state-of-the-art baselines. Code is available in the public repository \href{https://github.com/wzhSteve/DCFA}{(https://github.com/wzhSteve/DCFA)}.
\end{abstract}

\begin{figure}[t]
    \centering
    \includegraphics[width=0.48\textwidth]{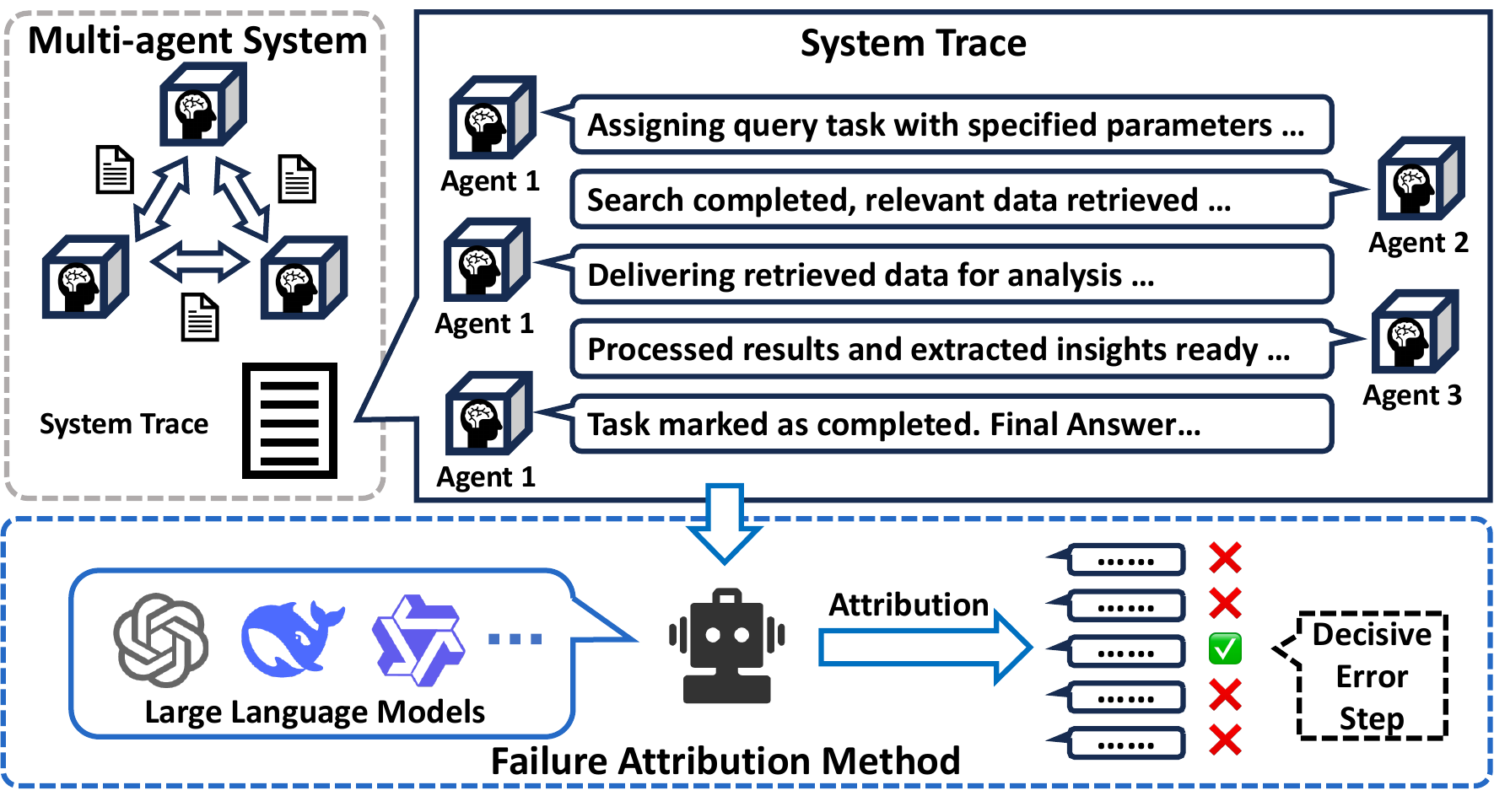}
    \caption{Illustration of failure attribution in an LLM-based multi-agent system.
    The system trace is represented as a textual record composed of interactive messages generated by agents. 
    }
    \label{fig:MAS_example}
\end{figure}

\section{Introduction}

With the emergence of large language models (LLMs), multi-agent systems (MAS) built upon LLM backends have rapidly evolved~\cite{becattini2025sallma,ronanki2025facilitating}. These systems enable autonomous collaboration among specialized agents for tasks such as code generation, knowledge search, and data analysis~\cite{huang2024romas,pan2024building,li2024survey,majdoub2025towards}. 
Despite the impressive progress in orchestration and tool integration, LLM-based multi-agent systems remain susceptible to coordination and reasoning failures~\cite{cemri2025multi,hammond2025multi,shen2025metacognitive}. Errors such as misalignment in agents' collaborations or improper tool usage can propagate across different agents along the system trace, leading to cascading reasoning errors that ultimately derail entire workflows~\cite{cemri2025multi,he2025sentinelagent}. 
This fragility highlights the urgent need for failure attribution in LLM-based multi-agent systems~\cite{cemri2025multi,hammond2025multi}, which refers to the process of identifying the earliest actions whose correction could reverse system failure, known as decisive errors.

Recent efforts have begun to explore failure attribution in LLM-based multi-agent systems using LLMs as diagnostic detectors~\cite{zhang2025agent,zhang2025agentracer}. Existing approaches can be broadly classified into fine-tuning-based~\cite{zhang2025agentracer} and instruction-based~\cite{zhang2025agent,cemri2025multi} paradigms. 
The fine-tuning-based approach, exemplified by AgenTracer~\cite{zhang2025agentracer}, constructs specialized datasets to train dedicated models for failure analysis. In contrast, instruction-based methods~\cite{zhang2025agent,cemri2025multi} rely on carefully designed prompts to guide LLMs in analyzing multi-agent system traces and identifying the decisive errors. 
However, both paradigms face inherent limitations.
Fine-tuning-based approaches offer strong controllability but require extensive annotation and repeated training, resulting in high costs.
Instruction-based methods avoid these costs under a training-free constraint by leveraging LLMs’ general reasoning ability, yet they often fail to reliably interpret long, multi-step, and unstructured system traces, leading to inaccurate failure attribution.

In this work, we study failure attribution in LLM-based multi-agent systems under the training-free setting to improve the extraction of decisive errors from unstructured multi-step traces.
We identify two fundamental challenges:
1) Shallow attribution:
Existing LLM-based attribution methods often focus on minor, recoverable deviations (e.g., incomplete retrieval or formatting errors~\cite{cemri2025multi,banerjee2025did}) that do not determine the final outcome. 
In contrast, the decisive error refers to the earliest action whose correction would reverse the system failure, truly driving the system to fail. 
For example, an agent may fabricate missing content when critical information is lost during the initial collection or retrieval process.
Unlike missing information that can be subsequently verified and retrieved through downstream interactions, the fabrication error is difficult to verify and thus persists throughout the reasoning chain, misleading the system’s subsequent decisions.
2) Contextual degradation:
As trace length and complexity increase, attribution performance degrades sharply.
For instance, the Who\&When benchmark~\cite{zhang2025agent} reports that the step-level attribution accuracy of state-of-the-art methods decreases substantially with growing trace length with averaging around 15\% for traces of length 10, but dropping to nearly 0\% for those approaching length 100.

To overcome these challenges, we propose a training-free framework, named the Dual-view Causal-inspired Failure Attribution (DCFA). DCFA comprises two key modules: the Global Causal-inspired Attribution (GCA) module and the Local Counterfactual-inspired Enhancement (LCE) module, which jointly perform global reasoning over a causal-inspired dependency structure and local counterfactual-inspired validation.
To move attribution beyond surface-level deviations, GCA constructs a structured causal-inspired dependency graph and performs reasoning along it to uncover deeper dependency relationships underlying system failure.
GCA extracts candidate minor deviations from the interaction system trace and builds a step-level causal-inspired dependency graph centered on these candidates. By performing global reasoning over the causal-inspired dependency graph across the entire system trace, GCA generates an initial hypothesis of the decisive error.
To mitigate long-context degradation, LCE refines the global hypothesis through localized counterfactual-inspired reasoning. Starting from the hypothesized error, LCE performs a bidirectional search over the causal-inspired dependency graph to identify local dependency chains through which the error propagates, and evaluates how local corrections propagate through the reasoning chain and affect the final outcome via counterfactual-inspired evaluation. By evaluating these locally induced outcome changes, LCE progressively identifies the interaction that most plausibly constitutes the decisive error.
In summary, our contributions are fourfold:
\begin{itemize}[nosep]
    \item We propose DCFA, a training-free framework for failure attribution in LLM-based multi-agent systems, integrating global reasoning with local counterfactual-inspired validation.
    \item To address shallow attribution, we construct causal-inspired dependency graphs to model interaction-level dependencies, enabling LLMs to move beyond minor deviations and identify decisive errors.
    \item To mitigate contextual degradation, we introduce localized counterfactual-inspired refinement that focuses reasoning on critical dependency chains, thereby improving attribution accuracy on long traces.
    \item Experiments on the Who\&When benchmark with six LLMs show consistent gains, improving step-level accuracy by up to 8.27\% over the state-of-the-art baselines.
\end{itemize}

\section{Related Works}

\subsection{LLM-based Multi-Agent Systems and System Failure}

Recent advances in LLM-based agents have driven rapid progress in multi-agent systems (MAS), enabling complex, long-horizon tasks through task decomposition and inter-agent communication~\cite{yang2025agentnet,sun2025llm}. Prior work has proposed systematic taxonomies of agent architectures~\cite{AG2_2024,fourney2024magentic}, identifying components such as role assignment, planning, external memory, and feedback mechanisms, which aim to bolster the problem-solving capabilities of LLM-based multi-agent systems.

Despite their promise, LLM-based multi-agent systems remain brittle~\cite{zhang2025agent,cemri2025multi}. Empirical studies reveal frequent coordination failures and cascading reasoning errors, particularly in long or highly interactive execution traces~\cite{liu2023agentbench}. These observations highlight the need for methods to analyze the decisive error of system failure.

\subsection{Failure Attribution in LLM-based Multi-agent Systems}

{
Failure attribution in multi-agent systems aims to identify the earliest mistake whose correction could reverse the overall system failure, referred to as the decisive error~\cite{cemri2025multi,fourney2024magentic}. Unlike prior work such as AgentErrorBench with AgentDebug~\cite{zhu2025llm}, which locates mistakes that trigger cascading reasoning distortions, or TRAIL~\cite{deshpande2025trail}, which identifies and analyzes all errors in trajectories, focusing on the decisive error directly targets the most actionable point for repairing task failure.
}

Recent approaches employ LLMs directly as diagnostic tools, including fine-tuning-based methods~\cite{zhang2025agentracer} and instruction-based methods~\cite{zhang2025agent,cemri2025multi}. Fine-tuned models offer strong control but require costly annotation and retraining. Instruction-based methods instead analyze execution traces via prompting. For instance, ECHO~\cite{banerjee2025did} is equipped with hierarchical context representations and multi-perspective consensus to improve attribution performance, while A2P~\cite{west2025abduct} is designed to perform causal-guided attribution through prompt-based instruction.

Despite their effectiveness, existing methods rely heavily on LLM intrinsic reasoning, often resulting in shallow attribution and degraded performance on long traces. To address these limitations, we propose a unified framework that combines structured reasoning over a causal-inspired dependency graph with localized dependency-chain analysis to improve failure attribution.

\subsection{Causal Reasoning and Counterfactual Analysis with LLMs}

Recent work has explored LLMs for causal reasoning and counterfactual-inspired analysis, showing that they can extract events, propose candidate causal relations, and simulate hypothetical interventions from unstructured text~\cite{cheng2025survey,luo2024open}. To improve robustness, several studies combine LLM-derived causal hypotheses with algorithmic validation or graph-based aggregation~\cite{tong2024automating,ban2025integrating}. Others treat LLMs as informative priors for causal-inspired dependency graph discovery, leveraging their encoded commonsense knowledge to guide structure learning~\cite{darvariu2024large,jiralerspong2024efficient}. Counterfactual simulation has also been used to probe the sensitivity and faithfulness of LLM reasoning chains~\cite{tutek2025measuring,yu2024causaleval}. These advances motivate our use of causal-inspired scaffolding and localized counterfactual-inspired validation for failure attribution in LLM-based multi-agent systems. Discussions of related work are deferred to Appendix~\ref{app:related_works}.

\section{Problem Formulation}

To respond to a user query, a multi-agent system generates a system trace including a sequence of interactions denoted as
$\mathcal{T} = \{\tau_1, \tau_2, \dots, \tau_n\}$, where each interaction
$\tau_t = (a_t, c_t)$ consists of the agent identifier $a_t$ and its corresponding content $c_t$.
The MAS produces a final output $\widehat{Y}$, which is expected to match the ground-truth outcome $Y$.
A system failure occurs when the generated output deviates from the expected one, i.e.,
$\widehat{Y} \neq Y$.
Such failures may arise from one or multiple erroneous or misleading interactions within $\mathcal{T}$.

The objective of failure attribution is to identify the decisive error responsible for the system failure.
When a single erroneous interaction accounts for the failure, the decisive error corresponds to the interaction whose correction can recover the correct outcome.
When multiple errors jointly contribute to the failure and no single correction fully recovers the correct outcome, we define the decisive error as the interaction whose correction yields the largest reduction in the system failure.

Formally, a failure attribution model $\mathcal{M}$ is defined as:
\begin{equation}
\mathcal{M}: (\mathcal{T}, \widehat{Y}) \mapsto \tau^{*},
\end{equation}
where $\tau^{*}$ denotes the ground-truth decisive error, i.e., the interaction whose correction either recovers the correct outcome when a single-step correction suffices, or maximally reduces system failure when multiple errors jointly contribute to the failure.

\begin{figure*}[htbp]
    \centering
    \includegraphics[width=0.98\textwidth]{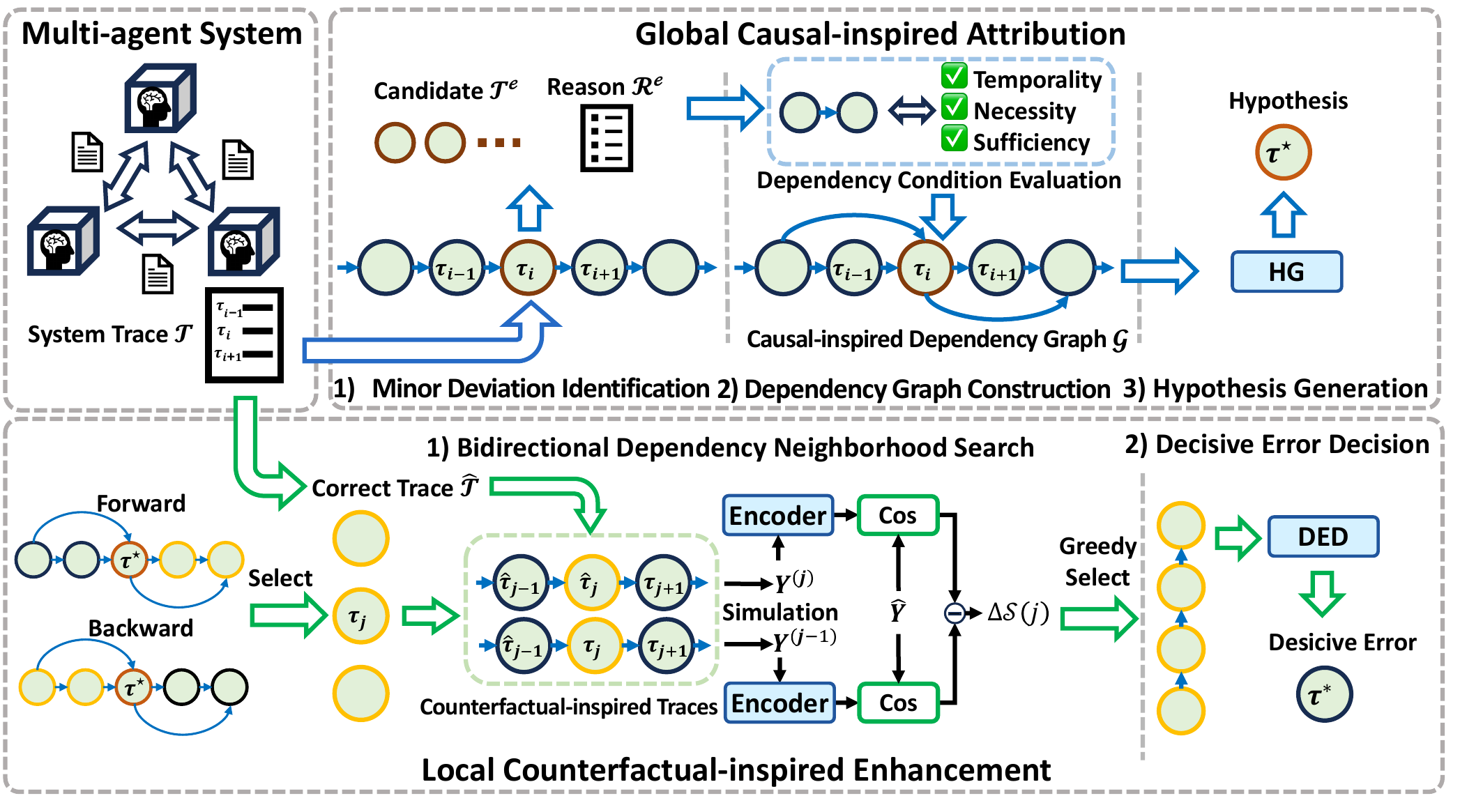}
    \caption{
    Overview of DCFA. DCFA integrates global causal-inspired attribution and local counterfactual-inspired enhancement to identify and refine the decisive error by reasoning over a causal-inspired dependency graph constructed from the system trace.}
    \label{fig:overview}
\end{figure*}

\section{Methodology}
\label{sec:methodology}

In this section, we introduce the proposed DCFA framework (Fig.~\ref{fig:overview}), which consists of the GCA (Sec.~\ref{sec:GCA}) and LCE (Sec.~\ref{sec:LCE}) modules.
To overcome shallow attribution, GCA reasons over full system traces to uncover latent dependency relationships, constructing a causal-inspired dependency graph and identifying an initial hypothesis of the decisive error from a global perspective.
Then, to further mitigate contextual degradation, the LCE module locally refines this hypothesis through counterfactual-inspired evaluation, using LLM-mediated approximate interventions to assess how correcting candidate interactions may alter the final outcome.
This strategy first establishes a global causal-inspired dependency structure and then locally validates its critical links, enabling more interpretable failure attribution for unstructured and long system traces.

\subsection{Global Causal-inspired Attribution Module}
\label{sec:GCA}

Interactions in system traces influence system outcomes through chained dependencies, requiring attribution methods that capture both minor deviations and how their effects propagate through the trace. Directly analyzing unstructured traces, LLMs often overemphasize salient deviations without sufficiently considering their downstream dependencies and influence on the final outcome.
To address this, GCA detects candidate deviations, constructs a directed causal-inspired dependency graph that links them to the full trace, and leverages graph-aware global reasoning to generate an initial hypothesis of the decisive error.

\subsubsection{Minor Deviation Identification}

Minor deviations are directly observable but typically benign faults in system traces, such as malformed outputs or incomplete information retrieval~\cite{cemri2025multi,hammond2025multi}. These deviations are often recoverable through built-in verification, tool re-invocation~\cite{barta2025measuring,zhou2025shielda}. However, minor deviations are not necessarily decisive errors, where they may serve as early symptoms of deeper failures.

Based on this motivation, GCA first extracts a set of candidate minor deviations that represent plausible manifestations of incorrect behaviors. Rather than prematurely determining which interaction constitutes the decisive error, GCA aims to retain all potentially relevant interactions for subsequent structured reasoning and refinement over the causal-inspired dependency graph.

Specifically, a designed process $\text{MDI}(\cdot)$ is utilized to detect the interactions of minor deviations and produce short natural-language explanations for each identified minor deviation:
\begin{equation}
(\mathcal{T}^{e}, \mathcal{R}^{e})  =  \text{MDI}\big(\mathcal{T}\big),
\end{equation}
where $\mathcal{T}^{e} = \{\tau^e_{1}, \tau^e_{2}, \dots\}$ is the set of candidate interactions that are identified as minor deviations,  $\mathcal{R}^{e}=\{r_1^{e},r_2^{e},\dots\}$ are the textual reasons associated with each candidate. 

\subsubsection{Dependency Graph Construction}

Having identified the candidate minor deviations and their corresponding textual reasons, GCA constructs a structured causal-inspired dependency graph $\mathcal{G}$ that encodes plausible dependency relationships among interactions in the system trace $\mathcal{T}$. The construction is centered on the detected minor deviations $\mathcal{T}^{e}$ and their associated reasoning cues $\mathcal{R}^{e}$, ensuring that the resulting structure reflects interpretable, error-centric reasoning pathways.

Formally, the causal-inspired dependency graph is denoted as $\mathcal{G} = (\mathcal{V}, \mathcal{E})$, where $\mathcal{V}$ is the set of step indicators $t$ corresponding to all interactions, and $\mathcal{E}$ is the set of directed edges representing putative dependency relationships. Specifically, an edge $e \in \mathcal{E}$ is denoted as $(p,q)$, indicating a directed dependency from interaction $\tau_p$ to $\tau_q$, where $\tau_p$ precedes and potentially influences $\tau_q$.

\paragraph{Dependency Condition Evaluation.}
A directed edge $e: \tau_p \rightarrow \tau_q$ is admitted into $\mathcal{E}$ only if it satisfies a set of structured dependency criteria~\cite{zeng2025zero,goldberg2019book}. For each ordered pair $(\tau_p,\tau_q)$ with step indices $p<q$, GCA evaluates three complementary properties: temporality, necessity, and sufficiency.

Temporality enforces the directional ordering constraint that a preceding interaction must occur before the downstream interaction:
\begin{equation}
\mathrm{Temp}(\tau_p \rightarrow \tau_q)=\mathbb{I}[p<q].
\end{equation}

Necessity tests whether the $\tau_q$ would fail to occur if the upstream step $\tau_p$ is removed, using a counterfactual-inspired judgment function $J$:
\begin{equation}
\mathrm{Nec}(\tau_p \rightarrow \tau_q)
=\mathbb{I}\!\left(J(\mathrm{do}(\neg\tau_p))=\text{No}\right).
\end{equation}

Sufficiency assesses whether enforcing $\tau_p$ is sufficient to plausibly induce the outcome at $\tau_q$:
\begin{equation}
\mathrm{Suff}(\tau_p \rightarrow \tau_q)
=\mathbb{I}\!\left(J(\mathrm{do}(\tau_p))=\text{Yes}\right).
\end{equation}

An edge is added only when the temporal, necessity, and sufficiency conditions are all satisfied. This conservative rule filters out spurious correlations, retaining causally supported dependencies.

\paragraph{Causal-inspired Structured Reasoning.}
Given the structured dependency criteria, GCA performs causal-inspired structured reasoning over the system trace $\mathcal{T}$, focusing on the identified minor deviations $\tau^{e}\in\mathcal{T}^{e}$ to construct the causal-inspired dependency graph $\mathcal{G}$.
The graph is constructed by incorporating these criteria into instruction sets $\mathcal{I}^{c}$:
\begin{equation}
\mathcal{G} = \mathrm{CGC}\big(\mathcal{T}, \mathcal{T}^{e}, \mathcal{R}^{e}, \mathcal{I}^{c}\big),
\label{eq:graph_generation}
\end{equation}
where $\mathrm{CGC}(\cdot)$ denotes the dependency graph construction procedure.

\subsubsection{Hypothesis Generation}

Based on the causal-inspired dependency graph $\mathcal{G}$ and the candidate minor deviation set $\mathcal{T}^{e}$, GCA performs a global analysis to propose an initial hypothesis of the decisive error of the system failure. 
The hypothesis is produced by the hypothesis generation process $\text{HG}(\cdot)$ on a compact evidence bundle that includes: the ground truth outcome $\widehat{\mathcal{Y}}$, the entire system trace $\mathcal{T}$, the candidate minor deviations $\mathcal{T}^{e}$, the corresponding textual reasons $\mathcal{R}^{e}$ to $\mathcal{T}^{e}$, and the constructed graph $\mathcal{G}$. 
Formally, the hypothesis generation is formulated as follows:
\begin{equation}
(\tau^{\star}, r^{\star}) =  \text{HG}\Big(\widehat{\mathcal{Y}},\,\mathcal{T},\,\mathcal{T}^{e},\,\mathcal{R}^{e},\,\mathcal{G}\Big),
\label{eq:hypothesis}
\end{equation}
where the output $\tau^{\star}$ is the hypothesis of the decisive error, and $r^{\star}$ is the corresponding reason.

\subsection{Local Counterfactual-inspired Enhancement Module}
\label{sec:LCE}

Practical system traces are often long and dense~\cite{zhang2025agent}, which exacerbates performance degradation in failure attribution based on full traces.
To address this limitation, we introduce the LCE module, which refines $\tau^{\star}$ by focusing on a compact local dependency neighborhood. LCE performs a bidirectional search over the causal-inspired dependency graph and uses counterfactual-inspired evaluation with LLM-mediated approximate interventions to assess how correcting candidate interactions may alter the final outcome. By iteratively evaluating local outcome changes, LCE identifies the interaction that most plausibly constitutes the decisive error.

\subsubsection{Bidirectional Dependency Neighborhood Search}

The bidirectional search aims to identify a minimal set of interactions that provide relevant evidence for explaining the system failure and to localize the decisive error by tracing dependency propagation across upstream and downstream interactions.
Starting from the index $i^{\star}$ of the global hypothesis $\tau^{\star}$, LCE initializes the pivot as $p^{(0)}=i^{\star}$ and explores the causal-inspired dependency graph $\mathcal{G}$ in both forward and backward directions.

The forward search traces downstream dependencies to capture how the hypothesized error may propagate, while the backward search examines upstream dependencies to identify prior misreasoning or missing premises. In both directions, interactions are selected greedily based on counterfactual-inspired evaluation, which assesses how modifying a candidate interaction may affect the final outcome. The union of the forward and backward results forms a locally refined dependency subchain centered on $\tau^{\star}$.

\paragraph{Counterfactual-inspired Evaluation}

Central to LCE is a counterfactual-inspired evaluation function that assesses the potential causal importance of an interaction by measuring the change in outcome alignment when that interaction is hypothetically corrected instead.

We first construct a fully corrected trace by applying a correction function $\text{Correct}(\cdot)$ conditioned on the ground-truth outcome $\widehat{Y}$, the original trace $\mathcal{T}$, extracted erroneous interactions $\mathcal{T}^{e}$, their corresponding reasons $\mathcal{R}^{e}$, and the causal-inspired dependency graph $\mathcal{G}$:
\begin{equation}
\widehat{\mathcal{T}} = \text{Correct}\left(\widehat{Y}, \mathcal{T}, \mathcal{T}^{e}, \mathcal{R}^{e}, \mathcal{G}\right),
\end{equation}
where $\widehat{\mathcal{T}} = \{\widehat{\tau}_1,\dots,\widehat{\tau}_N\}$ and $\widehat{\tau}_i$ denotes the corrected version of $\tau_i$.

Using $\widehat{\mathcal{T}}$, we construct a counterfactual-inspired trace for each interaction $\tau_i$ by combining the corrected prefix up to step $i$ with the original suffix from step $i\!+\!1$, which is denoted as $\widetilde{\mathcal{T}}^{(i)} = \{\widehat{\tau}_1,\dots,\widehat{\tau}_i,\tau_{i+1},\dots,\tau_N\}$.
This represents a scenario in which only the first $i$ interactions have been corrected. The counterfactual-inspired trace is then evaluated via a simulation process:
\begin{equation}
Y^{(i)} = \text{Simulation}\bigl(\widetilde{\mathcal{T}}^{(i)}\bigr).
\end{equation}

We measure the semantic alignment between the outcome $Y^{(i)}$ and the ground truth $\widehat{Y}$ using a pretrained encoder $\mathrm{Enc}(\cdot)$ and cosine similarity:
\begin{equation}
\mathcal{S}(\tau_i) = \mathrm{Cos}\bigl(\mathrm{Enc}(Y^{(i)}), \mathrm{Enc}(\widehat{Y})\bigr).
\end{equation}

The marginal attributional impact of correcting $\tau_i$ is defined as
\begin{equation}
\Delta \mathcal{S}(\tau_i) = \mathcal{S}(\tau_i) - \mathcal{S}(\tau_{i-1}),
\end{equation}
which measures the incremental improvement in outcome alignment obtained by correcting the $i$-th interaction after all preceding interactions have been corrected.

\paragraph{Bidirectional Greedy Search}

Using the counterfactual-inspired scores, LCE performs a greedy bidirectional search over the causal-inspired dependency graph $\mathcal{G}$. 

In the forward search, given the pivot $p^{(l-1)}$ at iteration $l$, we collect all direct successors denoted as $\mathcal{T}^{(l-1)}_{\text{eff}} = \{\tau_j \mid (p^{(l-1)}, j) \in \mathcal{E}\}$,
where $\mathcal{E}$ denotes the edge set of $\mathcal{G}$. The successor with the largest positive marginal contribution is selected:
\begin{equation}
\tau_{j^{(l)}} = \arg\max_{\tau_j \in \mathcal{T}^{(l-1)}_{\text{eff}}} \Delta \mathcal{S}(\tau_j).
\end{equation}

If $\Delta \mathcal{S}(\tau_{j^{(l)}}) > 0$, the forward candidate set, denoted as $\mathcal{C}_{\text{fwd}}$, is expanded and the pivot updated as $j^{(l)}$. Otherwise, the forward search terminates.

Symmetrically, the backward search traces incoming edges to identify upstream causes denoted as $\mathcal{T}^{(l-1)}_{\text{cau}} = \{\tau_k \mid (k, p^{(l-1)}) \in \mathcal{E}\}$.
The predecessor with the strongest positive effect is selected:
\begin{equation}
\tau_{k^{(l)}} = \arg\max_{\tau_k \in \mathcal{T}^{(l-1)}_{\text{cau}}} \Delta \mathcal{S}(\tau_k).
\end{equation}
If $\Delta \mathcal{S}(\tau_{k^{(l)}}) > 0$, the backward candidate set, denoted as $\mathcal{C}_{\text{bwd}}$, is expanded and the pivot updated as $k^{(l)}$. This process continues until no further positive contributions are found.

These two searches yield a local dependency neighborhood that captures both antecedent causes and propagated effects surrounding $\tau^{\star}$.

\subsubsection{Decisive Error Decision}

The forward and backward candidate sets are merged with the global hypothesis to form the final local candidate set $\mathcal{T}^{c} = \{\tau_i \mid i \in \{i^{\star}\} \cup \mathcal{C}_{\text{fwd}} \cup \mathcal{C}_{\text{bwd}}\}$.
Each element in $\mathcal{T}^{c}$ corresponds to an interaction whose counterfactual-inspired correction yields a positive improvement in outcome alignment.

Unlike GCA, which analyzes the full trace, LCE operates only on the interaction subset $\mathcal{T}^{c}$, avoiding context degradation. This evaluation is formalized by a decisive-error decision function:
\begin{equation}
\tau^{*} = \text{DED}\bigl(\widehat{Y}, \mathcal{T}^{c},\mathcal{G}\bigr),
\end{equation}
where $\tau^{*}$ denotes the final identified decisive error.

\section{Experimental Design}

\subsection{Dataset}
We evaluate DCFA on the Who\&When benchmark~\cite{zhang2025agent}, which is currently the only widely adopted benchmark specifically designed for failure attribution in LLM-based multi-agent systems and serves as the common evaluation testbed for existing baselines~\cite{west2025abduct,banerjee2025did}.
The dataset contains 184 multi-agent system traces spanning both Algorithm-generated and Hand-crafted settings. Specifically, 126 traces are generated using CaptainAgent-based systems~\cite{AG2_2024}, and 58 traces are manually curated from systems such as Magnetic-One~\cite{fourney2024magentic}. 
The tasks cover web-navigation and general reasoning scenarios derived from AssistantBench~\cite{yoran2024assistantbenchwebagentssolve} and GAIA~\cite{mialon2023gaia}. Each trace is annotated with fine-grained failure labels, including the decisive error and a natural-language explanation. Details are provided in Appendix~\ref{app:datasets}.

\subsection{Baselines}

We compare DCFA against five representative baselines that reflect different LLM-based failure attribution paradigms: All-at-Once, Step-by-Step, and Binary-Search from the Who\&When benchmark~\cite{zhang2025agent}, as well as A2P~\cite{west2025abduct} and ECHO~\cite{banerjee2025did}. All methods rely on LLMs for attribution and are evaluated under identical inference settings. We test both open-source and commercial models, including Qwen3-Coder-30B, Qwen3-235B, DeepSeek-R1-32B, DeepSeek-R1-671B, GPT-5, and Gemini-2.5-pro. Details are deferred to Appendix~\ref{app:baselines}.

\begin{table*}[!htbp]
\setlength{\tabcolsep}{2pt}
\centering
\scalebox{0.69}{
\begin{tabular}{l|cccccc|cccccc}
\toprule[1.5pt]
\multirow{2}{*}{Model} 
& \multicolumn{6}{c|}{Algorithm-generated} 
& \multicolumn{6}{c}{Hand-crafted} \\
\cline{2-13}
& All-at-Once & Step-by-Step & Binary-Search & A2P & ECHO & DCFA 
& All-at-Once & Step-by-Step & Binary-Search & A2P & ECHO & DCFA \\
\midrule[1.2pt]

Qwen3-Coder-30B & 11.90 & 20.63 & 13.49 & 12.70 & \underline{23.02} & \textbf{37.30} & 1.72 & 10.34 & 6.67 & 8.62 & \underline{13.79} & \textbf{15.52} \\

DeepSeek-R1-32B & 22.22 & 29.37 & 30.16 & 19.84 & \underline{30.95} & \textbf{42.06} & 5.17 & 15.52 & 6.90 & 6.90 & \underline{18.97} & \textbf{20.69} \\

Qwen3-235B & 30.95 & 25.40 & 21.43 & 29.37 & \underline{36.50} & \textbf{43.65} & 3.45 & 15.52 & 12.07 & 3.45 & \underline{17.24} & \textbf{22.41} \\

DeepSeek-R1-671B & 16.67 & 26.98 & 32.54 & 15.87 & \underline{40.48} & \textbf{53.17} & 3.45 & 12.07 & 5.17 & 6.90 & \underline{20.69} & \textbf{22.41} \\

GPT-5 & 15.87 & 27.78 & \underline{31.75} & 15.87 & \underline{31.75} & \textbf{47.62} & 3.45 & 13.79 & 10.34 & 13.79 & \underline{15.52} & \textbf{24.14} \\

Gemini-2.5-pro & 26.98 & \underline{34.13} & 25.40 & 32.54 & 30.95 & \textbf{46.03} & 5.17 & \underline{15.52} & 6.90 & 10.34 & 13.79 & \textbf{18.97} \\
\midrule[1.2pt]
Average & 20.77 & 27.72 & 25.13 & 21.03 & \underline{32.28} & \textbf{44.79} & 3.74 & 13.46 & 8.33 & 8.33 & \underline{16.67} & \textbf{20.69} \\
\bottomrule[1.5pt]
\end{tabular}
}
\caption{Step-level Accuracy Comparison on Algorithm-generated and Hand-crafted datasets.}
\label{tab:main_result_stepacc}
\end{table*}

\subsection{Implementation Details}

In DCFA, the GCA stage is performed by commercial LLMs using the same configurations as the baselines. The LCE stage is executed with a locally deployed Qwen-Coder-30B model.
The judgment function $J$ used for dependency condition evaluation in GCA is implemented via structured prompting, where the LLM is queried to perform binary judgments.
All LLMs are run with the temperature set to 0 to ensure deterministic inference. All baselines follow identical preprocessing pipelines and prompt configurations as specified in their original papers and released codebases to ensure fair comparison. As ECHO does not provide an official implementation, we reimplement it based on the descriptions in its paper.

We evaluate attribution quality using Step-level Accuracy, which directly measures precise failure localization and avoids the inflation effects of agent-level metrics. Additional implementation and evaluation details are provided in Appendix~\ref{app:implementation}.

\subsection{Results and Analysis}

\subsubsection{Overall Performance}

Table~\ref{tab:main_result_stepacc} reports step-level accuracy of DCFA and five baselines on the Algorithm-generated and Hand-crafted datasets. Compared with the strongest baseline on each dataset, DCFA achieves average improvements of 12.51\% and 4.02\%, respectively. 
On the Algorithm-generated dataset, DeepSeek-R1 achieves the highest accuracy, likely benefiting from reasoning-oriented training that aligns with DCFA’s structured causal-inspired dependency graph construction~\cite{guo2025deepseek}, while GPT remains competitive on longer traces due to its strong long-context reasoning~\cite{leon2025gpt}.
In contrast, most baselines rely primarily on prompt-induced implicit reasoning, making their attributions less sensitive to the latent dependency relations in multi-agent interactions. For instance, ECHO focuses on observable errors such as tool failures or formatting issues. While these explicit patterns can improve detection accuracy, the resulting deviations are often minor rather than decisive. By tracing such surface symptoms back to their underlying causes, DCFA enables more precise and causally grounded failure attribution.
Examples are presented in the case study (Sec.~\ref{sec:case_study}), with additional analysis of GCA-induced causal-inspired dependency graphs in Appendix~\ref{sec:causal_graph_example}.

\begin{table}[!htbp]
\setlength{\tabcolsep}{2pt}
\centering
\scalebox{0.76}{
\begin{tabular}{l|cc|cc}
\toprule[1.5pt]
\multirow{2}{*}{Model} 
& \multicolumn{2}{c|}{Algorithm-generated} 
& \multicolumn{2}{c}{Hand-crafted} \\
\cline{2-5}
& w/o LCE & DCFA & w/o LCE & DCFA \\
\midrule[1.2pt]
Qwen3-Coder-30B & 33.33 & \textbf{37.30} & 12.07 & \textbf{15.52}\\
DeepSeek-R1-32B & 40.48 & \textbf{42.06} & 15.52 & \textbf{20.69} \\
Qwen3-235B & 42.06 & \textbf{43.65} & 20.69 & \textbf{22.41} \\
DeepSeek-R1-671B & \textbf{53.17} & \textbf{53.17} & 18.97 & \textbf{22.41} \\
GPT-5 & 44.44 & \textbf{47.62} & 22.41 & \textbf{24.14} \\
Gemini-2.5-pro & 43.65 & \textbf{46.03} & 13.79 & \textbf{18.97} \\
\bottomrule[1.5pt]
\end{tabular}
}
\caption{Ablation study of the LCE module.}
\label{tab:ablation_study_stepacc_rel_gain}
\end{table}

\subsubsection{Ablation on LCE}
Since LCE operates on the causal-inspired dependency graph and the hypothesis of the decisive error produced by GCA, we compare GCA-only with DCFA.
Table~\ref{tab:ablation_study_stepacc_rel_gain} reports an ablation study of DCFA with and without LCE. Adding LCE consistently improves step-level accuracy on both datasets, with 3.45\% average improvement on the Hand-crafted dataset and 1.93\% average improvement on the Algorithm-generated dataset.

Notably, GCA alone exhibits a noticeable performance drop on the Hand-crafted dataset, where traces are substantially longer (average length $=51.6$) than those in the Algorithm-generated dataset (average length $=8.7$). 
In contrast, LCE remains effective even when global attribution degrades under long-context conditions. This robustness arises because LCE identifies the local dependency chains with the greatest corrective impact through bidirectional search and counterfactual-inspired evaluation, capturing both upstream causes and downstream effects of the error localized by GCA. By focusing the LLM on higher-level and more specific causal-inspired structures, LCE mitigates context degradation and improves attribution precision.

\subsubsection{Performance on Varying Trace Lengths}
\label{sec:varying_lengths}

We evaluate DCFA on the Hand-crafted dataset using DeepSeek-R1-671B and 32B, comparing it with all baselines across five context-length levels defined by the Who\&When benchmark~\cite{zhang2025agent}. 

\begin{figure}[t]
    \centering
    \subfloat[DeepSeek-R1-671B]{
    \includegraphics[width=0.223\textwidth]{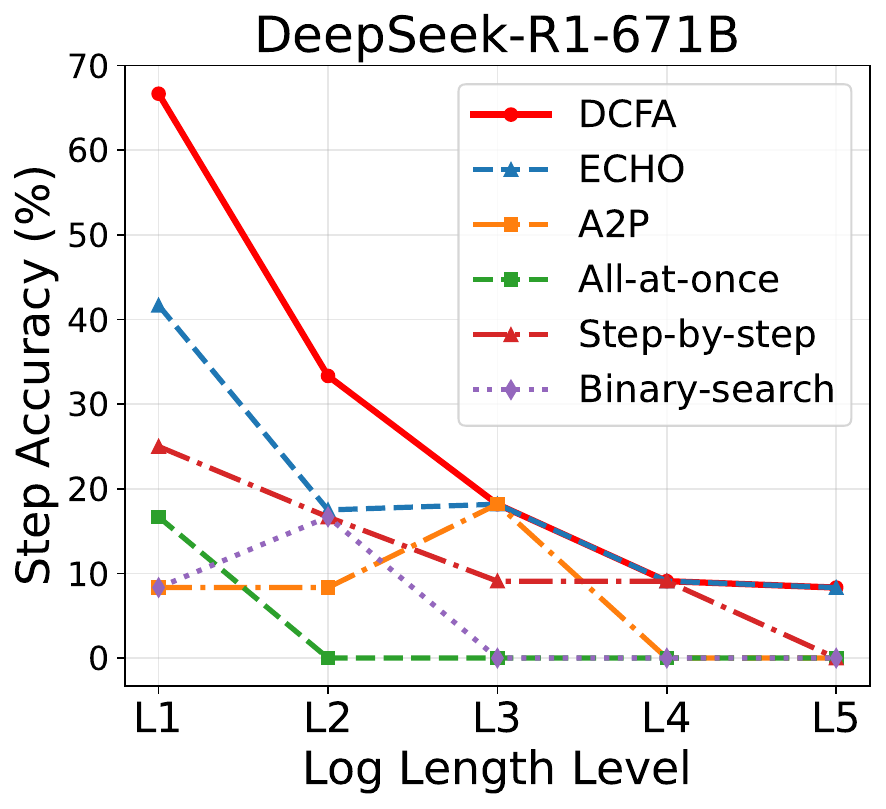}
    \label{fig:deepseek_671b}
    }
    \subfloat[DeepSeek-R1-32B]{
    \includegraphics[width=0.223\textwidth]{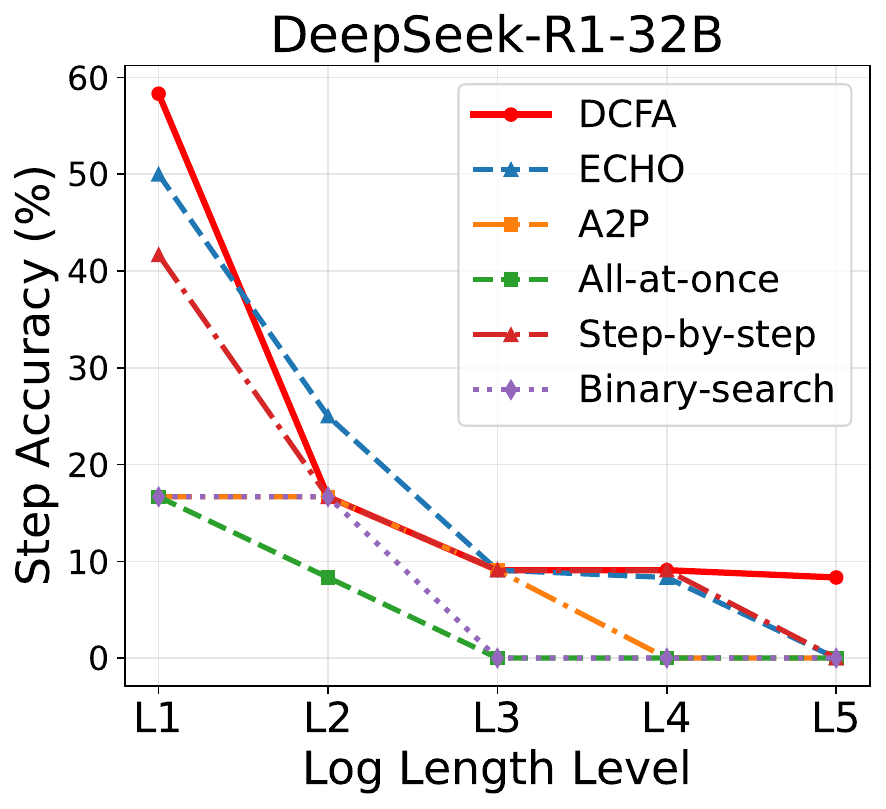}
    \label{fig:deepseek_32b}
    }
    \caption{Step-level accuracy of DCFA and baselines across trace lengths using DeepSeek-R1 series LLMs.}
    \label{fig:varying_trace}
\end{figure}

As shown in Fig.~\ref{fig:deepseek_671b} and Fig.~\ref{fig:deepseek_32b}, DCFA outperforms all baselines across nearly all levels, with the largest gains observed at Level~1 and maintained through Level~5. One exception occurs at Level~2 with DeepSeek-R1-32B, where DCFA slightly underperforms ECHO. This may be due to the reduced reliability of weaker LLMs in following the structured dependency reasoning required by DCFA, which can introduce additional variance in LLM-mediated evaluation. Nevertheless, DCFA remains robust overall, achieving strong performance across varying trace lengths.

\subsubsection{Case Study on Causal-inspired Dependency Graph Search}
\label{sec:case_study}

We analyze a system trace from a hand-crafted multi-agent system (ID: 49.json), where three agents ({Orchestrator}, {WebSurfer}, and {Assistant}) collaboratively solve an Unlambda debugging query (see details in Appendix~\ref{app:ex:user_query}). The execution trace $\mathcal{T} = \{\tau_1, \dots, \tau_{15}\}$ results in an incorrect prediction of ``k'', whereas the correct missing character is the backtick ``\`{}''.

\begin{figure}[htbp]
    \centering
    \includegraphics[width=0.45\textwidth]{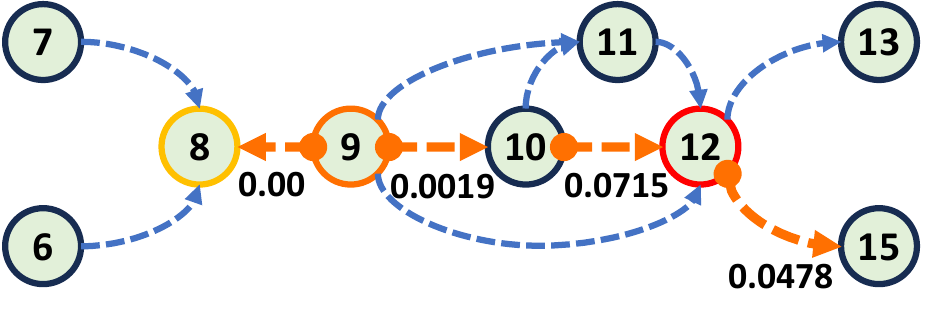}
    \caption{LCE refinement on the causal-inspired dependency graph of trace 49.json. Orange dashed arrows indicate search paths. Forward and backward traversal evaluates the corrective effects of candidate interactions, isolating $\tau_{12}$ as the decisive error. Yellow: minor deviation ($\tau_8$); Orange: GCA-identified error ($\tau_9$); Red: decisive error ($\tau_{12}$).}
    \label{fig:causal_graph}
\end{figure}

\paragraph{Inference Phase of GCA.}
GCA first identifies multiple erroneous or misleading interactions that are potentially involved in the failure. Specifically, it detects a minor deviation at $\tau_8$, where {WebSurfer} returns incomplete operator definitions, and further identifies $\tau_9$, where the {Orchestrator} propagates the partial information without further validation. These errors form a dependency chain and jointly contribute to the subsequent failure, rather than constituting independent failure sources. Based on the resulting causal-inspired dependency graph, GCA localizes the relevant error region for further refinement, as shown in Figure~\ref{fig:causal_graph}.

\paragraph{Refinement Phase via LCE.}
LCE then conducts localized, bidirectional counterfactual-inspired reasoning over the error region identified by GCA. By examining the corrective effects of candidate interactions along the dependency graph, LCE determines that $\tau_{12}$ has the greatest impact on mitigating the system failure. At $\tau_{12}$, the {Assistant} fabricates a solution based on incomplete semantics, which further propagates the accumulated errors toward the incorrect final outcome. Although correcting $\tau_8$ or $\tau_9$ can mitigate the error propagation, neither correction is sufficient to fully resolve the failure. In contrast, correcting $\tau_{12}$ provides the largest reduction in the failure, thereby identifying it as the decisive error. The refinement process is illustrated in Figure~\ref{fig:causal_graph}.

\paragraph{Decisive Error and DCFA Attribution.}
This case exemplifies the multi-error setting in our problem formulation, where multiple errors jointly contribute to the system failure while no single correction among the earlier errors fully recovers the correct outcome. Specifically, $\tau_8$ and $\tau_9$ contribute to the error propagation and can be corrected to mitigate the failure, whereas $\tau_{12}$ has the greatest corrective effect and is therefore identified as the decisive error. By combining global causal-inspired dependency search with local counterfactual-inspired refinement, DCFA successfully distinguishes contributing errors from the decisive error and accurately attributes the system failure to $\tau_{12}$. Additional details are provided in Appendix~\ref{app:case}.

\subsubsection{Computational Cost}

We analyze DCFA’s computational cost by decomposing it into GCA and LCE. 
GCA relies on commercial LLM API calls, while LCE is executed locally and adds only inference-time overhead. Following~\cite{zhang2025agent}, output-token costs are ignored unless stated otherwise.

GCA performs two full-context LLM calls: deviation detection with dependency graph construction, and decisive-error refinement. Since outputs are small relative to the input context, the total cost can be approximated as
\begin{equation}
\text{Cost}_{\text{GCA}} \approx 2C + 2\overline{n}L,
\end{equation}
where $C$ is the prompt overhead, $\overline{n}$ the average trace length, and $L$ the token length per step. Notably, the strongest baseline ECHO requires the same number of full-context calls, resulting in identical computational complexity, while GCA alone already yields substantial accuracy improvements.

The LCE module runs locally and introduces additional inference overhead. Its complexity can be approximated as
\begin{equation}
\text{Cost}_{\text{LCE}} \approx mC + m\overline{n}L,
\end{equation}
where $m \approx 4.5$ is the average number of bidirectional neighborhood expansions. Although LCE increases runtime, it further improves attribution accuracy, especially on longer trajectories. 

Table~\ref{tab:cost_comparison} summarizes the approximate runtime, token consumption, and attribution accuracy of all compared methods on the Who\&When benchmark. DCFA without LCE incurs a cost comparable to ECHO while achieving substantially higher accuracy ($30.05$ vs.\ $24.48$). Adding LCE increases the runtime from 32s to 300s and token consumption from 13k to 32k, but further improves accuracy to $32.74$. Detailed runtime statistics, token usage, and comprehensive comparisons with baseline methods are provided in Appendix~\ref{app:cost}.

\begin{table}[!t]
\centering
\scalebox{0.8}{
\begin{tabular}{lccc}
\toprule[1.5pt]
Method & Runtime (s) & Tokens (k) & Accuracy \\
\midrule[1.5pt]
All-at-Once & 15  & 6  & 12.26 \\
Search-by-Step & 61  & 9  & 20.59 \\
Binary-Search & 25  & 12 & 16.73 \\
A2P & 15  & 6  & 14.68 \\
ECHO & 31  & 12 & 24.48 \\
DCFA w/o LCE & 32 & 13 & \underline{30.05} \\
DCFA & 300 & 32 & \textbf{32.74} \\
\bottomrule[1.5pt]
\end{tabular}
}
\caption{Approximate runtime, token usage, and attribution accuracy on the Who\&When benchmark.}
\label{tab:cost_comparison}
\end{table}

\section{Conclusion}
In this work, we propose DCFA, a training-free framework for failure attribution in LLM-based multi-agent systems. Combining global causal-inspired dependency graph analysis with local counterfactual-inspired reasoning, DCFA identifies decisive errors and performs robustly across traces of varying lengths. 
Experiments show it improves step-level attribution accuracy by up to 8.27\% over state-of-the-art baselines and remains effective on challenging long-context traces.

\section*{Limitations}

\paragraph{Dependence on LLM Capability.}
DCFA relies on the reasoning inference capabilities of the underlying large language models. When system traces become very long, LLMs may struggle to maintain coherent dependency representations, which can reduce the accuracy of global causal-inspired dependency graph construction. Similarly, when using smaller or less capable models for local counterfactual-inspired refinement, the precision of decisive error identification may be constrained.

\paragraph{Scalability to Long or Complex Traces.}
Although the global-local combination in DCFA improves robustness to moderately long execution traces, extremely long or highly branching multi-agent interactions can still pose challenges. In such cases, both global causal-inspired attribution and local counterfactual-inspired reasoning may become less reliable, limiting applicability to systems with deeply nested or prolonged interaction patterns.

\paragraph{Attribution of Intrinsic LLM Reasoning Errors.}
Compared with the decisive errors that reflect failures within the MAS workflow. Errors originating from inherent LLM hallucinations may be difficult to attribute accurately.

\section*{Ethical Considerations}

This research is intended as a diagnostic tool to support failure analysis in LLM-based multi-agent systems, rather than an automated mechanism for judgment or accountability. Its attribution results depend on the reasoning behavior of underlying language models and may be imperfect, especially when failures stem from intrinsic model errors or ambiguous interactions. Over-reliance on such automated explanations could lead to misinterpretation of responsibility if used without human oversight. Therefore, we highlight that the proposed DCFA should be applied with caution in high-stakes settings and used to assist, not replace, human analysis of system failures.

\section*{Acknowledgment}
This work was supported in part by the National Natural Science Foundation of China under Grants 62572346 and 62322208.

\bibliography{anthology}

@article{deshpande2025trail,
  title={Trail: Trace reasoning and agentic issue localization},
  author={Deshpande, Darshan and Gangal, Varun and Mehta, Hersh and Krishnan, Jitin and Kannappan, Anand and Qian, Rebecca},
  journal={arXiv preprint arXiv:2505.08638},
  year={2025}
}

@article{zhu2025llm,
  title={Where llm agents fail and how they can learn from failures},
  author={Zhu, Kunlun and Liu, Zijia and Li, Bingxuan and Tian, Muxin and Yang, Yingxuan and Zhang, Jiaxun and Han, Pengrui and Xie, Qipeng and Cui, Fuyang and Zhang, Weijia and others},
  journal={arXiv preprint arXiv:2509.25370},
  year={2025}
}

@article{west2025abduct,
  title={Abduct, Act, Predict: Scaffolding Causal Inference for Automated Failure Attribution in Multi-Agent Systems},
  author={West, Alva and Weng, Yixuan and Zhu, Minjun and Lin, Zhen and Ning, Zhiyuan and Zhang, Yue},
  journal={arXiv preprint arXiv:2509.10401},
  year={2025}
}

@inproceedings{yu2024causaleval,
  title={Causaleval: Towards better causal reasoning in language models},
  author={Yu, Longxuan and Chen, Delin and Xiong, Siheng and Wu, Qingyang and Li, Dawei and Chen, Zhikai and Liu, Xiaoze and Pan, Liangming},
  booktitle={Proceedings of the 2025 Conference of the Nations of the Americas Chapter of the Association for Computational Linguistics: Human Language Technologies (Volume 1: Long Papers)},
  pages={12512--12540},
  year={2025}
}

@inproceedings{tutek2025measuring,
  title={Measuring chain of thought faithfulness by unlearning reasoning steps},
  author={Tutek, Martin and Chaleshtori, Fateme Hashemi and Marasovi{\'c}, Ana and Belinkov, Yonatan},
  booktitle={Proceedings of the 2025 Conference on Empirical Methods in Natural Language Processing},
  pages={9946--9971},
  year={2025}
}

@inproceedings{verma2025causal,
  title={Causal AI Scientist: Facilitating Causal Data Science with Large Language Models},
  author={Verma, Vishal and Acharya, Sawal and Simko, Samuel and Bhardwaj, Devansh and Haghighat, Anahita and Janzing, Dominik and Sachan, Mrinmaya and Jin, Zhijing and Yang, Yongjin},
  booktitle={NeurIPS 2025 AI for Science Workshop},
  year={2025}
}

@inproceedings{zhu2024causal,
  title={Causal inference with latent variables: Recent advances and future prospectives},
  author={Zhu, Yaochen and He, Yinhan and Ma, Jing and Hu, Mengxuan and Li, Sheng and Li, Jundong},
  booktitle={Proceedings of the 30th ACM SIGKDD Conference on Knowledge Discovery and Data Mining},
  pages={6677--6687},
  year={2024}
}

@article{liu2025large,
  title={Large language models and causal inference in collaboration: A comprehensive survey},
  author={Liu, Xiaoyu and Xu, Paiheng and Wu, Junda and Yuan, Jiaxin and Yang, Yifan and Zhou, Yuhang and Liu, Fuxiao and Guan, Tianrui and Wang, Haoliang and Yu, Tong and others},
  journal={Findings of the Association for Computational Linguistics: NAACL 2025},
  pages={7668--7684},
  year={2025}
}

@article{ban2025integrating,
  title={Integrating large language model for improved causal discovery},
  author={Ban, Taiyu and Chen, Lyuzhou and Lyu, Derui and Wang, Xiangyu and Zhu, Qinrui and Tu, Qiang and Chen, Huanhuan},
  journal={IEEE Transactions on Artificial Intelligence},
  year={2025},
  publisher={IEEE}
}

@article{cohrs2025large,
  title={Large language models for causal hypothesis generation in science},
  author={Cohrs, Kai-Hendrik and Diaz, Emiliano and Sitokonstantinou, Vasileios and Varando, Gherardo and Camps-Valls, Gustau},
  journal={Machine Learning: Science and Technology},
  volume={6},
  number={1},
  pages={013001},
  year={2025},
  publisher={IOP Publishing}
}

@article{tong2024automating,
  title={Automating psychological hypothesis generation with AI: when large language models meet causal graph},
  author={Tong, Song and Mao, Kai and Huang, Zhen and Zhao, Yukun and Peng, Kaiping},
  journal={Humanities and Social Sciences Communications},
  volume={11},
  number={1},
  pages={896},
  year={2024},
  publisher={Palgrave}
}

@inproceedings{luo2024open,
  title={Open event causality extraction by the assistance of llm in task annotation, dataset, and method},
  author={Luo, Kun and Zhou, Tong and Chen, Yubo and Zhao, Jun and Liu, Kang},
  booktitle={Proceedings of the Workshop: Bridging Neurons and Symbols for Natural Language Processing and Knowledge Graphs Reasoning (NeusymBridge)@ LREC-COLING-2024},
  pages={33--44},
  year={2024}
}

@article{cheng2025survey,
  title={A Survey of Event Causality Identification: Taxonomy, Challenges, Assessment, and Prospects},
  author={Cheng, Qing and Zeng, Zefan and Hu, Xingchen and Si, Yuehang and Liu, Zhong},
  journal={ACM Computing Surveys},
  volume={58},
  number={3},
  pages={1--37},
  year={2025},
  publisher={ACM New York, NY}
}

@inproceedings{su2025enhancing,
  title={Enhancing Event Causality Identification with LLM Knowledge and Concept-Level Event Relations},
  author={Su, Ya and Zhang, Hu and Zhang, Guangjun and Wang, Yujie and Fan, Yue and Li, Ru and Wang, Yuanlong},
  booktitle={Proceedings of the 31st International Conference on Computational Linguistics},
  pages={7403--7414},
  year={2025}
}

@misc{goldberg2019book,
  title={The Book of Why: The New Science of Cause and Effect: by Judea Pearl and Dana Mackenzie, Basic Books (2018). ISBN: 978-0465097609.},
  author={Goldberg, Lisa R},
  year={2019},
  publisher={Taylor \& Francis}
}

@article{he2025sentinelagent,
  title={SentinelAgent: Graph-based Anomaly Detection in Multi-Agent Systems},
  author={He, Xu and Wu, Di and Zhai, Yan and Sun, Kun},
  journal={arXiv preprint arXiv:2505.24201},
  year={2025}
}

@inproceedings{liu2023agentbench,
  title={Agentbench: Evaluating llms as agents},
  author={Liu, Xiao and Yu, Hao and Zhang, Hanchen and Xu, Yifan and Lei, Xuanyu and Lai, Hanyu and Gu, Yu and Ding, Hangliang and Men, Kaiwen and Yang, Kejuan and others},
  booktitle={International Conference on Learning Representations},
  volume={2024},
  pages={52989--53046},
  year={2024}
}

@article{yang2025agentnet,
  title={Agentnet: Decentralized evolutionary coordination for llm-based multi-agent systems},
  author={Yang, Yingxuan and Chai, Huacan and Shao, Shuai and Song, Yuanyi and Qi, Siyuan and Rui, Renting and Zhang, Weinan},
  journal={Advances in Neural Information Processing Systems},
  volume={38},
  pages={107309--107336},
  year={2026}
}

@article{shrotriya2025navigating,
  title={Navigating the path to Ethical \& Responsible AI integration in Health \& Life Sciences with Human and Machine Collaboration},
  author={Shrotriya, Shobhit and Banu, N and Kulkarni, Avi and Aiyangar, Sujata},
  journal={Journal of the Society for Clinical Data Management},
  volume={5},
  number={1},
  pages={1--12},
  year={2025},
  publisher={Society for Clinical Data Management}
}

@inproceedings{li2022training,
  title={Training data debugging for the fairness of machine learning software},
  author={Li, Yanhui and Meng, Linghan and Chen, Lin and Yu, Li and Wu, Di and Zhou, Yuming and Xu, Baowen},
  booktitle={Proceedings of the 44th International Conference on Software Engineering},
  pages={2215--2227},
  year={2022}
}

@article{zhou2025shielda,
  title={SHIELDA: Structured Handling of Exceptions in LLM-Driven Agentic Workflows},
  author={Zhou, Jingwen and Chen, Jieshan and Lu, Qinghua and Zhao, Dehai and Zhu, Liming},
  journal={arXiv preprint arXiv:2508.07935},
  year={2025}
}

@incollection{barta2025measuring,
  title={Measuring the Robustness of Multi-Agent Reinforcement Learning Systems under Partial Agent Failure},
  author={Barta, Zolt{\'a}n and Nagy, Bal{\'a}zs and Guly{\'a}s, L{\'a}szl{\'o}},
  booktitle={Proceedings of the Intelligent Robotics FAIR 2025},
  pages={58--63},
  year={2025}
}

@article{zeng2025zero,
  title={Zero-shot event causality identification via multisource evidence fuzzy aggregation with large language models},
  author={Zeng, Zefan and Cheng, Qing and Hu, Xingchen and Li, Wentao and Ding, Weiping and Liu, Zhong},
  journal={IEEE Transactions on Fuzzy Systems},
  year={2026},
  publisher={IEEE}
}

@inproceedings{pan2024building,
  title={Building Multi-Agent Copilot towards Autonomous Agricultural Data Management and Analysis},
  author={Pan, Yu and Sun, Jianxin and Yu, Hongfeng and Luck, Joe and Bai, Geng and Chamara, Nipuna and Ge, Yufeng and Awada, Tala},
  booktitle={2024 IEEE International Conference on Big Data (BigData)},
  pages={4384--4393},
  year={2024},
  organization={IEEE}
}

@article{imai2024causal,
  title={Causal Inference with Generative Artificial Intelligence: Application to Texts as Treatments},
  author={Imai, Kosuke and Nakamura, Kentaro},
  journal={Journal of the American Statistical Association},
  number={just-accepted},
  pages={1--27},
  year={2026},
  publisher={Taylor \& Francis}
}

@inproceedings{jiralerspong2024efficient,
  title={Efficient Causal Graph Discovery Using Large Language Models},
  author={Jiralerspong, Thomas and Chen, Xiaoyin and More, Yash and Shah, Vedant and Bengio, Yoshua},
  booktitle={ICLR 2024 Workshop: How Far Are We From AGI}
}

@article{darvariu2024large,
  title={Large language models are effective priors for causal graph discovery},
  author={Darvariu, Victor-Alexandru and Hailes, Stephen and Musolesi, Mirco},
  journal={arXiv preprint arXiv:2405.13551},
  year={2024}
}

@inproceedings{wang2025event,
  title={Event causality identification with synthetic control},
  author={Wang, Haoyu and Liu, Fengze and Zhang, Jiayao and Roth, Dan and Richardson, Kyle},
  booktitle={Proceedings of the 2024 Conference on Empirical Methods in Natural Language Processing},
  pages={1725--1737},
  year={2024}
}

@inproceedings{liu2024identifying,
  title={Identifying while Learning for Document Event Causality Identification},
  author={Liu, Cheng and Xiang, Wei and Wang, Bang},
  booktitle={Proceedings of the 62nd Annual Meeting of the Association for Computational Linguistics (Volume 1: Long Papers)},
  pages={3815--3827},
  year={2024}
}

@inproceedings{shen2025metacognitive,
  title={Metacognitive self-correction for multi-agent system via prototype-guided next-execution reconstruction},
  author={Shen, Xu and Zhang, Qi and Wang, Song and Tan, Zhen and Zhao, Xinyu and Yao, Laura and Tadiparthi, Vaishnav and Mahjoub, Hossein Nourkhiz and Pari, Ehsan Moradi and Lee, Kwonjoon and others},
  booktitle={Findings of the Association for Computational Linguistics: ACL 2026},
  pages={23320--23337},
  year={2026}
}

@article{xia2025demystifying,
  title={Demystifying llm-based software engineering agents},
  author={Xia, Chunqiu Steven and Deng, Yinlin and Dunn, Soren and Zhang, Lingming},
  journal={Proceedings of the ACM on Software Engineering},
  volume={2},
  number={FSE},
  pages={801--824},
  year={2025},
  publisher={ACM New York, NY, USA}
}

@inproceedings{ronanki2025facilitating,
  title={Facilitating Trustworthy Human-Agent Collaboration in LLM-based Multi-Agent System oriented Software Engineering},
  author={Ronanki, Krishna},
  booktitle={Proceedings of the 33rd ACM International Conference on the Foundations of Software Engineering},
  pages={1333--1337},
  year={2025}
}

@inproceedings{becattini2025sallma,
  title={SALLMA: A Software Architecture for LLM-Based Multi-Agent Systems},
  author={Becattini, Marco and Verdecchia, Roberto and Vicario, Enrico},
  booktitle={2025 IEEE/ACM International Workshop New Trends in Software Architecture (SATrends)},
  pages={5--8},
  year={2025},
  organization={IEEE}
}

@inproceedings{pei2025flow,
  title={Flow-of-Action: SOP Enhanced LLM-Based Multi-Agent System for Root Cause Analysis},
  author={Pei, Changhua and Wang, Zexin and Liu, Fengrui and Li, Zeyan and Liu, Yang and He, Xiao and Kang, Rong and Zhang, Tieying and Chen, Jianjun and Li, Jianhui and others},
  booktitle={Companion Proceedings of the ACM on Web Conference 2025},
  pages={422--431},
  year={2025}
}

@inproceedings{majdoub2025towards,
  title={Towards Adaptive Software Agents for Debugging},
  author={Majdoub, Yacine and Charrada, Eya Ben and Touati, Haifa},
  booktitle={Proceedings of the 33rd ACM International Conference on the Foundations of Software Engineering},
  pages={636--640},
  year={2025}
}

@article{hammond2025multi,
  title={Multi-agent risks from advanced ai},
  author={Hammond, Lewis and Chan, Alan and Clifton, Jesse and Hoelscher-Obermaier, Jason and Khan, Akbir and McLean, Euan and Smith, Chandler and Barfuss, Wolfram and Foerster, Jakob and Gaven{\v{c}}iak, Tom{\'a}{\v{s}} and others},
  journal={arXiv preprint arXiv:2502.14143},
  year={2025}
}

@article{li2024survey,
  title={A survey on LLM-based multi-agent systems: workflow, infrastructure, and challenges},
  author={Li, Xinyi and Wang, Sai and Zeng, Siqi and Wu, Yu and Yang, Yi},
  journal={Vicinagearth},
  volume={1},
  number={1},
  pages={9},
  year={2024},
  publisher={Springer}
}

@article{huang2024romas,
  title={Romas: A role-based multi-agent system for database monitoring and planning},
  author={Huang, Yi and Cheng, Fangyin and Zhou, Fan and Li, Jiahui and Gong, Jian and Yang, Hongjun and Fan, Zhidong and Jiang, Caigao and Xue, Siqiao and Chen, Faqiang},
  journal={arXiv preprint arXiv:2412.13520},
  year={2024}
}

@article{sun2025llm,
  title={LLM-Based Multi-Agent Decision-Making: Challenges and Future Directions},
  author={Sun, Chuanneng and Huang, Songjun and Pompili, Dario},
  journal={IEEE Robotics and Automation Letters},
  year={2025},
  publisher={IEEE}
}

@inproceedings{zhang2025agentracer,
  title={Agentracer: Who is inducing failure in the llm agentic systems?},
  author={Zhang, Guibin and Wang, Junhao and Chen, Junjie and Zhou, Wangchunshu and Wang, Kun and Yan, Shuicheng},
  booktitle={International Conference on Learning Representations},
  volume={2026},
  pages={11377--11399},
  year={2026}
}

@article{AG2_2024,
  title={Autogen: Enabling next-gen llm applications via multi-agent conversation},
  author={Wu, Qingyun and Bansal, Gagan and Zhang, Jieyu and Wu, Yiran and Li, Beibin and Zhu, Erkang and Jiang, Li and Zhang, Xiaoyun and Zhang, Shaokun and Liu, Jiale and others},
  journal={arXiv preprint arXiv:2308.08155},
  year={2023}
}

@inproceedings{yoran2024assistantbenchwebagentssolve,
  title={Assistantbench: Can web agents solve realistic and time-consuming tasks?},
  author={Yoran, Ori and Amouyal, Samuel Joseph and Malaviya, Chaitanya and Bogin, Ben and Press, Ofir and Berant, Jonathan},
  booktitle={Proceedings of the 2024 Conference on Empirical Methods in Natural Language Processing},
  pages={8938--8968},
  year={2024}
}

@inproceedings{mialon2023gaia,
  title={Gaia: a benchmark for general ai assistants},
  author={Mialon, Gr{\'e}goire and Fourrier, Cl{\'e}mentine and Wolf, Thomas and LeCun, Yann and Scialom, Thomas},
  booktitle={The Twelfth International Conference on Learning Representations},
  year={2023}
}

@article{fourney2024magentic,
  title={Magentic-one: A generalist multi-agent system for solving complex tasks},
  author={Fourney, Adam and Bansal, Gagan and Mozannar, Hussein and Tan, Cheng and Salinas, Eduardo and Niedtner, Friederike and Proebsting, Grace and Bassman, Griffin and Gerrits, Jack and Alber, Jacob and others},
  journal={arXiv preprint arXiv:2411.04468},
  year={2024}
}

@InProceedings{zhang2025agent,
  title = 	 {Which Agent Causes Task Failures and When? {O}n Automated Failure Attribution of {LLM} Multi-Agent Systems},
  author =       {Zhang, Shaokun and Yin, Ming and Zhang, Jieyu and Liu, Jiale and Han, Zhiguang and Zhang, Jingyang and Li, Beibin and Wang, Chi and Wang, Huazheng and Chen, Yiran and Wu, Qingyun},
  booktitle = 	 {Proceedings of the 42nd International Conference on Machine Learning},
  pages = 	 {76583--76599},
  year = 	 {2025},
  volume = 	 {267},
  series = 	 {Proceedings of Machine Learning Research},
  month = 	 {13--19 Jul},
  publisher =    {PMLR},
}

@article{yan2025beyond,
  title={Beyond self-talk: A communication-centric survey of llm-based multi-agent systems},
  author={Yan, Bingyu and Zhou, Zhibo and Zhang, Litian and Zhang, Lian and Zhou, Ziyi and Miao, Dezhuang and Li, Zhoujun and Li, Chaozhuo and Zhang, Xiaoming},
  journal={arXiv preprint arXiv:2502.14321},
  year={2025}
}

@article{cemri2025multi,
  title={Why do multi-agent llm systems fail?},
  author={Cemri, Mert and Pan, Melissa Z and Yang, Shuyi and Agrawal, Lakshya A and Chopra, Bhavya and Tiwari, Rishabh and Keutzer, Kurt and Parameswaran, Aditya and Klein, Dan and Ramchandran, Kannan and others},
  journal={Advances in Neural Information Processing Systems},
  volume={38},
  year={2026}
}

@article{leon2025gpt,
  title={GPT-5 and open-weight large language models: Advances in reasoning, transparency, and control},
  author={Leon, Maikel},
  journal={Information Systems},
  pages={102620},
  year={2025},
  publisher={Elsevier}
}

@article{guo2025deepseek,
  title={Deepseek-r1: Incentivizing reasoning capability in llms via reinforcement learning},
  author={Guo, Daya and Yang, Dejian and Zhang, Haowei and Song, Junxiao and Wang, Peiyi and Zhu, Qihao and Xu, Runxin and Zhang, Ruoyu and Ma, Shirong and Bi, Xiao and others},
  journal={arXiv preprint arXiv:2501.12948},
  year={2025}
}

@inproceedings{banerjee2025did,
  title={Where Did It All Go Wrong? A Hierarchical Look into Multi-Agent Error Attribution},
  author={Banerjee, Adi and Nair, Anirudh and Borogovac, Tarik},
  booktitle={NeurIPS 2025 Workshop on Evaluating the Evolving LLM Lifecycle: Benchmarks, Emergent Abilities, and Scaling},
    year={2025},
  journal={Advances in Neural Information Processing Systems},
}

\appendix
\section{Dataset} \label{app:datasets}
We utilize the datasets from the Who\&When benchmark~\cite{zhang2025agent}, which comprise both Algorithm-generated and Hand-crafted multi-agent system traces, totaling 184 executions. 
Specifically, 126 traces are generated algorithmically using multi-agent systems built on CaptainAgent~\cite{AG2_2024}, while 58 traces are manually curated from Hand-crafted systems such as Magnetic-One~\cite{fourney2024magentic}. 
The dataset encompasses a wide range of realistic multi-agent scenarios based on queries from GAIA~\cite{mialon2023gaia} and AssistantBench~\cite{yoran2024assistantbenchwebagentssolve}.

The dataset covers a broad set of multi-agent tasks, including web-navigation-style challenges derived from AssistantBench~\cite{yoran2024assistantbenchwebagentssolve} and general-assistant reasoning tasks inspired by GAIA~\cite{mialon2023gaia}. 
Each trace is annotated with fine-grained failure information, including the responsible agent, the true decisive error, and a natural language explanation of the failure.

\begin{figure}[htbp]
    \centering
    \includegraphics[width=0.48\textwidth]{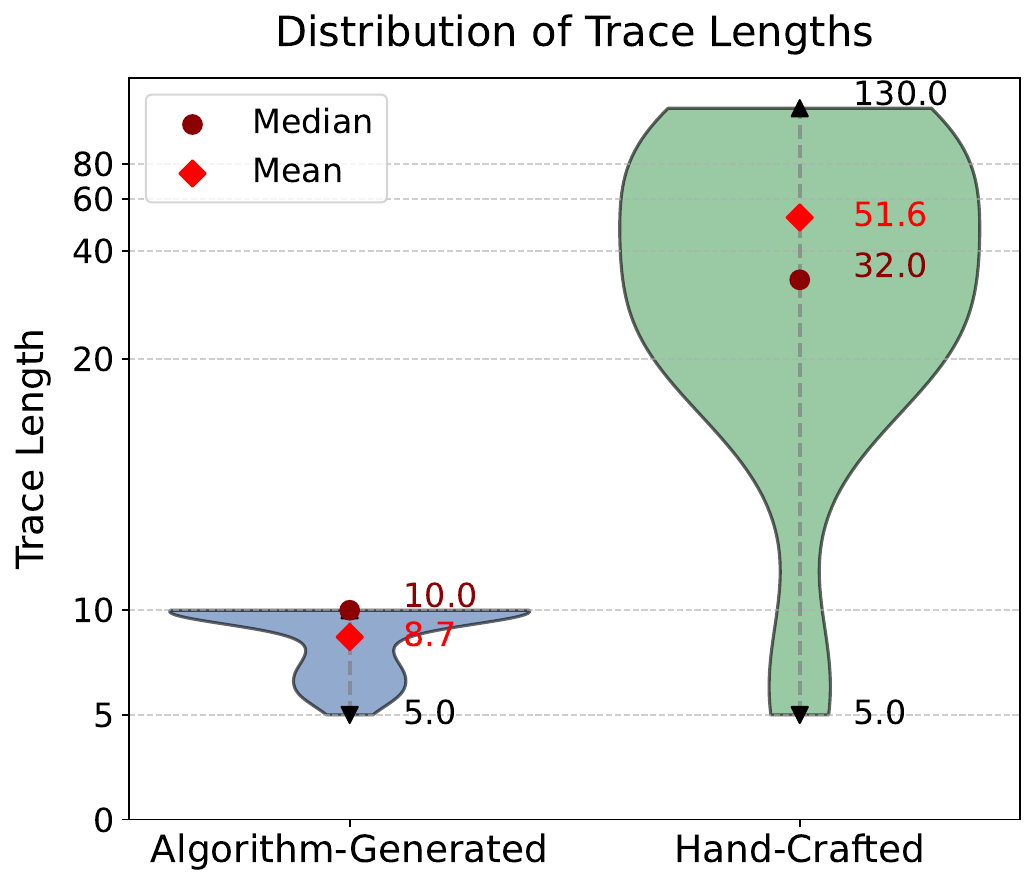}
    \caption{Violin plot showing the distribution of trace lengths for the Algorithm-generated and Hand-crafted datasets. The width of each violin represents the kernel density estimation of the data distribution, illustrating the frequency of different trace lengths. The central red line indicates the median, while the black triangles mark the minimum and maximum values. Notably, wider regions denote higher concentrations of trace lengths within that range.}
    \label{fig:dataset}
\end{figure}

The two categories exhibit distinct characteristics in terms of agent participation and trace length. Algorithm-generated traces involve less than 4 agents, with trace lengths ranging from 5 to 10 steps, a mean length of 8.7 steps, and a median length of 10 steps. In contrast, Hand-crafted traces involve between 1 and 5 agents, with trace lengths spanning 5 to 130 steps, a mean length of 51.6 steps, and a median length of 32 steps. The distributions of both datasets, illustrated in Fig.~\ref{fig:dataset}, highlight the greater variability present in the Hand-crafted traces.

{
Moreover, some benchmarks, such as AgentErrorBench with AgentDebug~\cite{zhu2025llm} and TRAIL~\cite{deshpande2025trail}, may appear superficially similar to our task setting. However they differ fundamentally from our problem formulation and evaluation objective, making direct comparison inappropriate. AgentErrorBench focuses on human-agent interaction and single-agent trajectories, where the objective is to detect the earliest mistake that initiates cascading reasoning distortions leading to failure. 
TRAIL is designed for identifying and analyzing all errors in trajectories. 
In contrast, this work focuses on failure attribution in LLM-based multi-agent systems, where the decisive error is defined as the earliest mistake whose correction could reverse the overall system failure. For example, an upstream deviation (an agent returning incomplete search results) eventually leads to a downstream decisive error (another agent fabricating missing information). AgentErrorBench would typically detect the earlier deviation, which could potentially be mitigated through subsequent agent interactions such as verification or additional tool usage, whereas this work aims to identify the interaction whose counterfactual-inspired correction would directly rescue the final outcome.
}

\section{Baselines} \label{app:baselines}

To evaluate the effectiveness of DCFA, we compare it against five representative baseline strategies that differ in how the LLM interacts with the failure traces: All-at-Once, Step-by-Step, and Binary-Search from the Who\&When benchmark~\cite{zhang2025agent}, as well as A2P~\cite{west2025abduct} and ECHO~\cite{banerjee2025did}. Each baseline reflects a distinct reasoning paradigm in localizing the failure-responsible step and agent within multi-agent system traces.

\begin{itemize}[nosep]
    \item \textbf{All-at-Once.}  
    The algorithm leverages the LLM to interpret the entire system trace in a single pass and directly infer the decisive error of the system failure.

    \item \textbf{Step-by-Step.}  
    The algorithm partitions the complete system trace into individual interactions. 
    At each step, the model determines whether an error has occurred in the current interaction. 
    If an error is detected, the process terminates immediately and the LLM outputs the corresponding step index. 
    Otherwise, the reasoning continues sequentially until the final interaction is reached.

    \item \textbf{Binary-Search.}  
    The algorithm begins with the query and the full failure log. 
    It first determines whether the fault occurred in the upper or lower half of the trace. 
    The identified half is then provided for further inspection. 
    This iterative procedure continues until a single interaction is isolated as the decisive error. 
    \item \textbf{A2P.}
    The algorithm employs a causal-inspired scaffold that sequentially guides LLMs through abductive hypothesis generation, explicit action-level intervention, and short-horizon counterfactual-inspired simulation to assess whether correcting a given step can reverse a system failure.

    \item \textbf{ECHO.} The algorithm introduces a hierarchical error attribution approach for multi-agent systems, explicitly targeting both minor execution deviations and overt errors within long interaction traces. Through multi-level contextual abstraction and consensus-based analysis, ECHO can detect early, subtle deviations that precede downstream failures, enabling fine-grained and reliable error localization.

\end{itemize}

All the baselines are based on the LLMs to achieve the failure attribution. To comprehensively assess each algorithm, we test them with various LLMs, including both open-source models of different sizes and commercial models. The models tested include the Qwen3 series (Qwen3-Coder-30B and Qwen3-235B), the DeepSeek series (DeepSeek-R1-32B and DeepSeek-R1-671B), GPT-5, and Gemini-2.5-pro. All baselines are evaluated under identical inference conditions to ensure a fair and consistent comparison.

\section{Implementation Details}
\label{app:implementation}

Our approach operates in a training-free manner. In the GCA stage, minor deviation identification, causal-inspired dependency graph construction, and hypothesis generation are performed by the selected commercial LLM, following the same configuration as the baseline setup. In the LCE stage, the bidirectional dependency neighborhood search and decisive error decision are executed using the locally deployed Qwen-Coder-30B model. 
The encoder used to transform the outcomes generated by the simulation process into a continuous embedding space is implemented using the text-embedding-3-large API provided by OpenAI. 

The judgment function $J$ is implemented as a prompt-based evaluator operating over multi-agent system traces. Rather than re-executing the entire system, $J$ receives a system trace containing agent identities and message contents during in the interactions, and is prompted to assess whether a target interaction would plausibly occur under a counterfactual-inspired intervention

Commercial LLMs are accessed through their respective APIs, while for local experiments, we deploy Qwen-Coder-30B on a workstation equipped with two NVIDIA A800 GPUs. For both commercial and locally deployed LLMs, the temperature is fixed to 0 to ensure deterministic outputs.

All inference procedures of the baseline methods follow identical preprocessing steps and prompt configurations on the Who\&When benchmark to ensure fair and reproducible evaluation across different models.

To evaluate attribution quality, we focus on Step-level Accuracy, which measures the proportion of cases in which the exact erroneous step is correctly identified. Notably, we do not focus on Agent-level Accuracy, as in systems with a small number of agents, Agent-level Accuracy can be inflated. This occurs because it overlooks the more precise Step-level localization of failures. Given our emphasis on accurate localization at both the agent and step levels, we exclusively use Step-level Accuracy, rejecting the misleading Agent-level baseline.

\section{Robustness Across Independent Runs}
\label{app:robustness}

Although we set the decoding temperature to 0 for all experiments, minor randomness in LLM inference may still lead to small fluctuations in the outputs. To evaluate the robustness of our results, we repeated the experiments using the three best-performing LLMs and report the mean and standard deviation over three independent runs.

As shown in Table~\ref{tab:robustness}, the variance across runs is extremely small for all methods and models. This indicates that the observed performance differences are stable and not caused by stochastic variations in the LLM outputs. In particular, DCFA consistently achieves the highest performance across all settings with minimal variance, demonstrating the robustness of the proposed failure attribution framework.

\begin{table*}[t]
\centering
\scalebox{0.78}{
\begin{tabular}{lccc|ccc}
\toprule
\multirow{2}{*}{Method} & \multicolumn{3}{c}{Algorithm-Generated} & \multicolumn{3}{c}{Hand-Crafted} \\
\cmidrule(lr){2-4} \cmidrule(lr){5-7}
 & DeepSeek-R1-671B & GPT-5 & Gemini-2.5-pro & DeepSeek-R1-671B & GPT-5 & Gemini-2.5-pro \\
\midrule
All-at-Once & 16.93 $\pm$ 0.00 & 16.14 $\pm$ 0.00 & 27.51 $\pm$ 0.02 & 4.02 $\pm$ 0.01 & 3.45 $\pm$ 0.00 & 4.60 $\pm$ 0.01 \\
Step-by-Step & 27.25 $\pm$ 0.03 & 28.31 $\pm$ 0.03 & 34.92 $\pm$ 0.03 & 11.50 $\pm$ 0.01 & 13.22 $\pm$ 0.01 & 16.10 $\pm$ 0.01 \\
Binary-Search & 33.86 $\pm$ 0.01 & 31.75 $\pm$ 0.03 & 26.46 $\pm$ 0.02 & 5.75 $\pm$ 0.01 & 10.92 $\pm$ 0.01 & 5.75 $\pm$ 0.01 \\
A2P & 16.40 $\pm$ 0.04 & 16.40 $\pm$ 0.05 & 31.75 $\pm$ 0.05 & 6.32 $\pm$ 0.01 & 12.64 $\pm$ 0.01 & 9.77 $\pm$ 0.01 \\
ECHO & 40.21 $\pm$ 0.00 & 31.48 $\pm$ 0.03 & 31.22 $\pm$ 0.00 & 19.54 $\pm$ 0.09 & 13.79 $\pm$ 0.01 & 13.22 $\pm$ 0.08 \\
DCFA & \textbf{53.44 $\pm$ 0.01} & \textbf{47.09 $\pm$ 0.04} & \textbf{46.69 $\pm$ 0.01} & \textbf{21.38 $\pm$ 0.04} & \textbf{22.41 $\pm$ 0.01} & \textbf{21.84 $\pm$ 0.03} \\
\bottomrule
\end{tabular}
}
\caption{Robustness evaluation across three independent runs. We report mean performance and standard deviation. The extremely small variance indicates that the results are stable despite potential randomness in LLM inference.}
\label{tab:robustness}
\end{table*}

\section{Detailed Computational Cost Analysis}
\label{app:cost}

\subsection{Computational Cost Calculation}

This section reports detailed empirical statistics of token usage and runtime for DCFA and baseline methods on the Who\&When benchmark. 
For DCFA, the average prompt overhead is approximately 400 tokens. The average trace length is about 3k tokens for the Algorithm-Generated dataset and 10k tokens for the Hand-Crafted dataset. 
The GCA stage performs two full-context API calls, each taking roughly 16 seconds, resulting in an average runtime of about 32 seconds per instance. 
The LCE stage is executed locally using Qwen3-Coder-30B on two NVIDIA A800 GPUs. Each inference processes approximately 6k tokens and takes about 60 seconds. 
Since bidirectional dependency neighborhood search requires on average 4.5 such calls, the total LCE runtime is approximately 270 seconds. 
Consequently, the overall runtime of DCFA is around 300 seconds per trajectory.

Baseline methods differ primarily in how they process the interaction trace. 
All-at-Once analyzes the entire trace using a single LLM call of roughly 6k tokens. 
Search-by-Step processes one interaction at a time (about 50 tokens per step) and averages around 20 steps per trace, resulting in approximately 9k tokens. 
Binary-Search iteratively analyzes halves of the remaining trace and typically requires about five calls, leading to roughly 12k tokens. 
A2P uses a prompt of about 300 tokens and processes the full trace in a single call, while ECHO also uses a prompt of about 300 tokens but performs two full-context calls.

\subsection{Trade-off Analysis}
Although DCFA introduces additional computational overhead, it yields substantial performance improvements. In terms of computational complexity, the strongest baseline, ECHO, and the GCA component of DCFA both require two full-context LLM calls.
Notably, the GCA module alone already achieves clear improvements over ECHO (+10.58\% on the Algorithm-Generated dataset and +0.57\% on the Hand-Crafted dataset), suggesting that the primary performance gains stem from enhanced reasoning over the trajectory.
The LCE module, in contrast, relies on local LLMs and therefore incurs relatively additional cost. It further improves detection accuracy on longer trajectories (e.g., +3.45\% on the Hand-Crafted dataset), indicating a favorable trade-off between attribution accuracy and computational efficiency.

\subsection{Runtime Cost and Practicality}
DCFA is designed as a post-hoc failure attribution framework, consistent with baselines such as ECHO and A2P, rather than as a real-time debugging module. Its primary goal is to analyze the complete system trace after a task failure and identify the decisive errors. Insights from this analysis can guide offline modifications to the MAS, including agent structures, context management strategies, or tool invocation policies, rather than adjusting the system during execution.

For online or real-time scenarios, a different design is required. Incremental reasoning can be applied at each interaction step, leveraging cached causal-inspired dependency subgraphs and restricting counterfactual-inspired evaluations to local neighborhoods. This avoids repeatedly processing the full trace and can substantially reduce runtime overhead, making failure attribution more practical for long-running or continuously operating multi-agent systems.

\begin{figure*}[!ht]
    \centering
    \subfloat[{Gemini-2.5-pro}]{
    \includegraphics[width=0.98\textwidth]{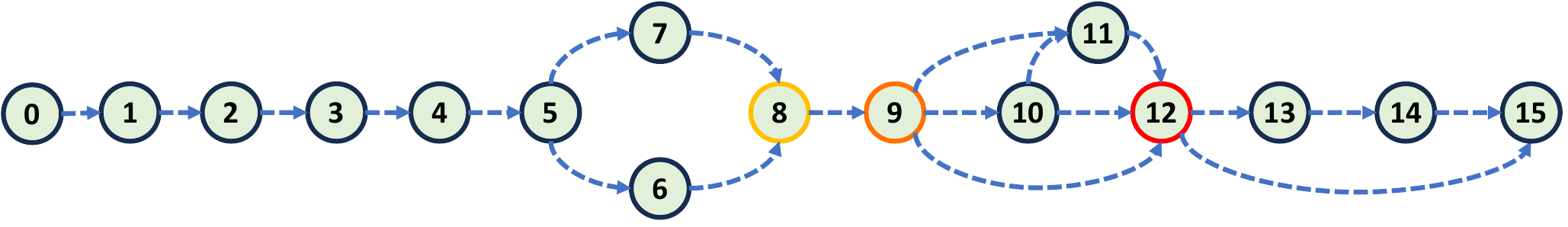}
    }\\
    \centering
    \subfloat[{GPT-5}]{
    \includegraphics[width=0.98\textwidth]{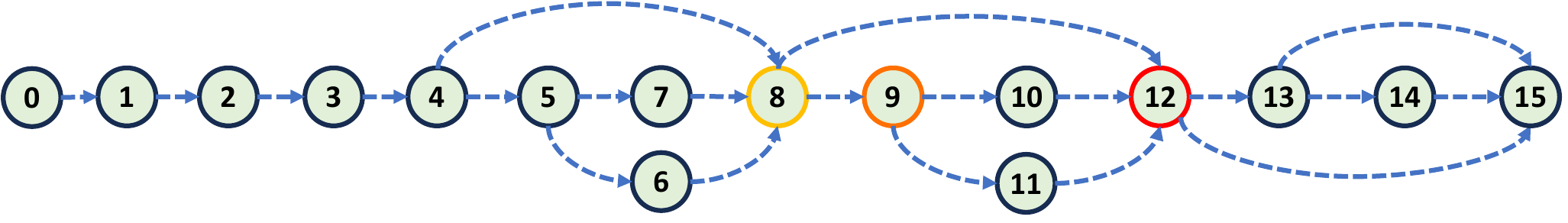}
    }
    \caption{Representative causal-inspired dependency graphs constructed by Gemini-2.5-pro and GPT-5 on the Hand-crafted example 49.json. Differences are most pronounced in the critical failure region $\tau_9$--$\tau_{12}$.}
    \label{fig:causal_graph_examples}
\end{figure*}

\section{Supplementary Analysis of GCA-Induced Causal-inspired Dependency Graphs}
\label{sec:causal_graph_example}

We analyze the structural statistics of the causal-inspired dependency graphs constructed during the GCA stage, with a particular focus on cross-model variability and its implications for decisive error localization.
To facilitate a concrete comparison, we further present representative causal-inspired dependency graphs generated by GPT-5 and Gemini-2.5-pro on a selected sample (49.json) from the Hand-crafted dataset.

\subsection{Model Capacity and Deviations in Causal-inspired Dependency Graph Edge Density}
Table~\ref{tab:edge_per_graph_stats} reports the average number of edges per graph across different LLMs on both the Algorithm-generated and Hand-crafted datasets.

\begin{table}[!htbp]
\setlength{\tabcolsep}{2pt}
\centering
\scalebox{0.8}{
\begin{tabular}{l|c|c}
\toprule[1.5pt]
Model & Algorithm-generated & Hand-crafted \\
\midrule[1.2pt]
Qwen3-Coder-30B & 7.89 & 68.90 \\
DeepSeek-R1-32B & 8.24 & 46.39 \\
Qwen3-235B & 9.41 & 61.52 \\
DeepSeek-R1-671B & 9.25 & 57.09 \\
GPT-5 & 10.12 & 55.40 \\
Gemini-2.5-pro & 8.76 & 56.38 \\
\midrule[1pt]
Average & 8.95 & 56.61 \\
\bottomrule[1.5pt]
\end{tabular}
}
\caption{Average number of edges per causal-inspired dependency graph generated during the GCA stage on the Algorithm-generated and Hand-crafted datasets.}
\label{tab:edge_per_graph_stats}
\end{table}

A clear stratification emerges when comparing models of different capability.
Less capable models, such as Qwen3-Coder-30B and DeepSeek-R1-32B, exhibit pronounced deviations from the overall mean, especially on the Hand-crafted dataset, producing overly dense or sparse causal-inspired dependency graphs (e.g., 68.90 edges for Qwen3-Coder-30B and 46.39 edges for DeepSeek-R1-32B).
Such extreme deviations, in either direction, indicate unstable causal induction and low quality of the causal-inspired dependency graph generation, likely due to context degradation in LLMs.

By contrast, higher-capability models, such as GPT-5, Gemini-2.5-pro, and DeepSeek-R1-671B, maintain edge counts closer to the global average, suggesting a stronger ability to suppress irrelevant dependencies and preserve salient relations.

\subsection{Implications for Decisive Error Localization.}

To qualitatively assess how these structural differences affect downstream reasoning, Figure~\ref{fig:causal_graph_examples} visualizes the causal-inspired dependency graphs constructed by Gemini-2.5-Pro and GPT-5 on the same hand-crafted example, 49.json.
Although both models successfully capture the overall dependency structure, their local structures in the critical failure region differ substantially.

Specifically, the GCA stage using Gemini-2.5-Pro initially hypothesizes $\tau_9$ as the decisive error, reflecting a more entangled dependency subgraph in the interval $\tau_9$--$\tau_{12}$.
In contrast, GCA with GPT-5 constructs a more compact and hierarchical causal-inspired dependency structure over the same region, enabling it to identify $\tau_{12}$ as the decisive error.

Importantly, this discrepancy does not prevent DCFA with Gemini-2.5-Pro from ultimately recovering the correct failure source.
Through the LCE module in the subsequent DCFA stage, the model performs counterfactual-inspired reasoning over the extracted local dependency subgraph, progressively evaluating and eliminating less relevant upstream candidates and converging on the decisive error $\tau_{12}$.
This observation highlights that while reliable global causal-inspired dependency graphs facilitate earlier localization, the proposed framework remains robust to structural noise through localized counterfactual-inspired refinement.

\section{Edge Growth and Long-Trace Scalability}

Analysis of the causal-inspired dependency graphs constructed by DCFA shows that edge growth is nearly linear with trajectory length. In realistic multi-agent traces, each interaction typically links only to temporally or semantically adjacent steps, rather than to all preceding interactions. For example, in the Algorithm-generated dataset, traces average 8.7 steps with 8.95 edges, while in the Hand-crafted dataset, traces average 51.6 steps with 56.61 edges. This correspondence indicates that edge growth scales linearly rather than quadratically, mitigating combinatorial complexity.

Maintaining consistent structured reasoning over long traces remains challenging for LLM-based approaches. LCE addresses this challenge by performing localized exploration within the dependency graph, reducing the reasoning scope and keeping the additional search overhead roughly linear in trace length. Alternative cost-efficient strategies could further alleviate scaling issues, including heuristic graph pruning (e.g., limiting search depth or filtering low-confidence edges), sliding-window reasoning without full graph traversal, or hierarchical trace compression prior to graph construction. These approaches offer options for scaling DCFA to longer or more complex trajectories while retaining the benefits of localized dependency reasoning.

\section{Details for Case Study on Causal-inspired Dependency Graph Search}
\label{app:case}

To demonstrate the effectiveness of DCFA, we present a case study based on a system trace from the Hand-crafted multi-agent system (ID: 49.json). In this scenario, the user query triggers interactions among three agents: Orchestrator, WebSurfer, and Assistant. The Orchestrator decomposes the task and coordinates the reasoning workflow, the WebSurfer retrieves external information, and the Assistant integrates the gathered evidence to produce the final answer.

\begin{figure*}[htbp]
    \centering
    \includegraphics[width=0.98\textwidth]{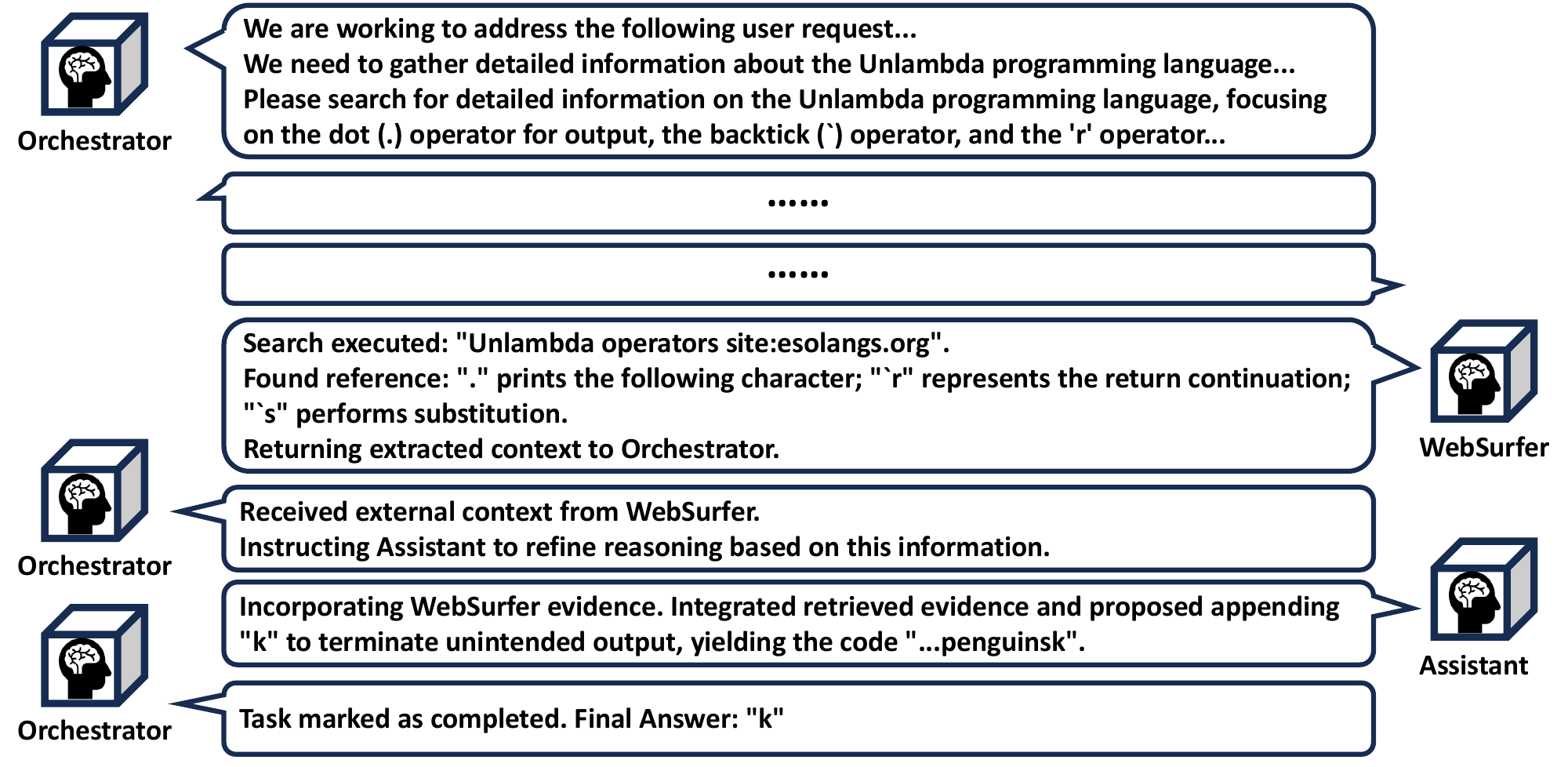}
    \caption{Key interaction steps of incorrect answer generation in trace 49.json.}
    \label{fig:example_case_study}
\end{figure*}

\begin{table*}[!htpb]
\centering
\scalebox{0.9}{
\begin{tabular}{lccc}
\toprule[1.5pt]
Stage & Identified Decisive Error  & Agent  & Supporting Evidence \\
\midrule[1.5pt]
GCA (Global) & $\tau_9$ & Orchestrator & Ignore missing operator details \\
LCE (Local) & $\tau_{12}$ & Assistant & Fabricate missing content \\
\bottomrule[1.5pt]
\end{tabular}
}
\caption{Comparison between GCA and LCE Attributions.}
\label{tab:compare}
\end{table*}

The system trace captures each step of the multi-agent reasoning process, including information collection, context propagation, and intermediate reasoning actions. 
Specifically, the detail content of the user query is shown in the following box:

\Examplebox{User Query}{
\label{app:ex:user_query}
In Unlambda, what exact character or text should be added to make the given code output ``For penguins''? Answer with the name of the needed character if applicable.\\
Code: `r```````````.F.o.r. .p.e.n.g.u.i.n.si
}
In response, the multi-agent system generates the system trace consisting of fifteen interactions $\mathcal{T} = \{\tau_{1}, \dots, \tau_{15}\}$. 
Each represents an interaction from one agent, as illustrated in Fig.~\ref{fig:example_case_study}. The system produces an incorrect answer ``k'', whereas the correct missing character should be  backtick ``\`{}''.

Within this trace, some steps involve minor deviations, such as incomplete retrieval or partial context propagation, which are generally correctable through downstream verification. In contrast, a decisive error occurs when the Assistant fabricates information based on incomplete or ambiguous input, producing content that is irreversible and misguides subsequent reasoning, ultimately leading to an incorrect outcome.

This overview sets the stage for a detailed case study analysis, where we examine how DCFA distinguishes contributory minor deviations from the decisive fabrication error and demonstrates its capability to pinpoint the root cause of failure in multi-agent reasoning.

\paragraph{Inference Phase of GCA}
During global analysis, GCA first detects a potential minor deviation at $\tau_8$, where the WebSurfer attempts to retrieve definitions of Unlambda operators but returns incomplete information.

\Examplebox{Excerpt from $\tau_8$}{
\label{app:ex:tau8}
WebSurfer: Searching for Unlambda operators ``.'', ``\`{}'', and ``r''...
}

Based on this, GCA builds a causal-inspired dependency graph centered on $\tau_8$ (Fig.~\ref{fig:causal_graph}) and ultimately identifies and attributes the system failure to $\tau_9$, where the Orchestrator mistakenly assumes that the retrieved information is complete and proceeds without verification.

\Examplebox{Excerpt from $\tau_{9}$}{
\label{app:ex:tau9}
Orchestrator: Progress confirmed. Proceeding to analyze gathered information...
}

\paragraph{Refinement Phase of LCE.}
To refine this hypothesis of GCA, the Local Counterfactual-inspired Enhancement (LCE) module focuses on the neighborhood around $\tau_9$, forming a local subgraph (Fig.~\ref{fig:causal_graph}). It conducts counterfactual-inspired simulations to evaluate whether modifying prior steps would correct the outcome. Forward traversal from $\tau_9$ to $\tau_8$ shows no effect, while backward traversal identifies $\tau_{12}$ as the decisive cause.

\Examplebox{Excerpt from $\tau_{12}$}{
\label{app:ex:tau12}
Assistant: Summarizing Unlambda constructs... The backtick (`) applies functions, the dot (.) outputs characters, and the ``r'' operator reads inputs. Based on this, adding ``k'' may terminate unwanted outputs.
}

LCE identifies $\tau_{12}$ as the decisive error, where the Assistant fabricates knowledge by falsely claiming that the WebSurfer’s incomplete results included operator definitions. This misinformation misleads subsequent reasoning, ultimately producing the wrong output in $\tau_{15}$.

\paragraph{Decisive Error and DCFA Attribution.}
Table~\ref{tab:compare} compares the decisive error attributions identified by the GCA and LCE modules. In this trace, $\tau_8$ corresponds to the WebSurfer attempting to retrieve essential information about Unlambda operators.
Although the retrieved information is incomplete, this step constitutes a minor deviation: it is benign and can be effectively corrected by the system’s subsequent verification mechanisms, such as detecting missing fields and re-invoking tools to retrieve the omitted information. Therefore, $\tau_8$ does not constitute a decisive error.

$\tau_9$ represents the Orchestrator proceeding and propagating the partial context returned by WebSurfer. While the GCA module flags $\tau_9$ as a global-level misjudgment, this step primarily propagates the minor deviation from $\tau_8$ rather than introducing a fundamentally new error, making it contributory but not causative.

The decisive error occurs at $\tau_{12}$, where the Assistant integrates the retrieved information and fabricates operator details, transforming a recoverable incompleteness into an irreversible factual error. This fabricated knowledge directly misguides subsequent reasoning and drives the final incorrect outcome. Unlike $\tau_8$, the error at $\tau_{12}$ cannot be corrected through downstream verification, highlighting its role as the decisive upstream failure.

The DCFA framework identifies this decisive error through a two-stage reasoning process. First, GCA constructs a step-level causal-inspired dependency graph centered on minor deviations, capturing high-level dependencies and propagating influence to generate an initial hypothesis, here pointing to $\tau_9$. Second, LCE performs a local neighborhood analysis around the hypothesis, employing counterfactual-inspired simulations to quantify the impact of each step on the final outcome. By evaluating the potential corrections in context, LCE isolates $\tau_{12}$ as the only step whose modification could prevent the error, thereby pinpointing the true decisive error.

This dual-view strategy, combining global reasoning over a causal-inspired dependency structure with local counterfactual-inspired validation, allows DCFA to transition from broad dependency mapping to precise error localization, effectively distinguishing contributory minor deviations from decisive errors. The approach not only enhances the interpretability of multi-agent reasoning traces but also identifies critical interactions whose correction is most likely to improve the final outcome, demonstrating clear advantages over purely global or local attribution methods.

\begin{figure*}[!htbp]
    \centering
    \subfloat[DeepSeek-R1-671B]{
    \includegraphics[width=0.98\textwidth]{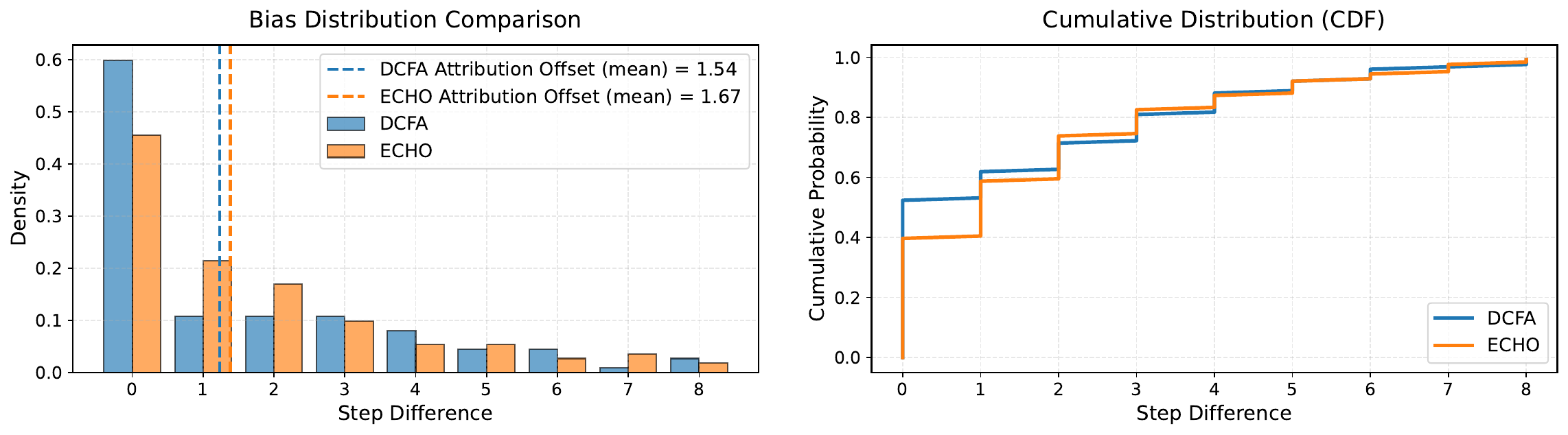}
    }\\
    \centering
    \subfloat[Gemini-2.5-pro]{
    \includegraphics[width=0.98\textwidth]{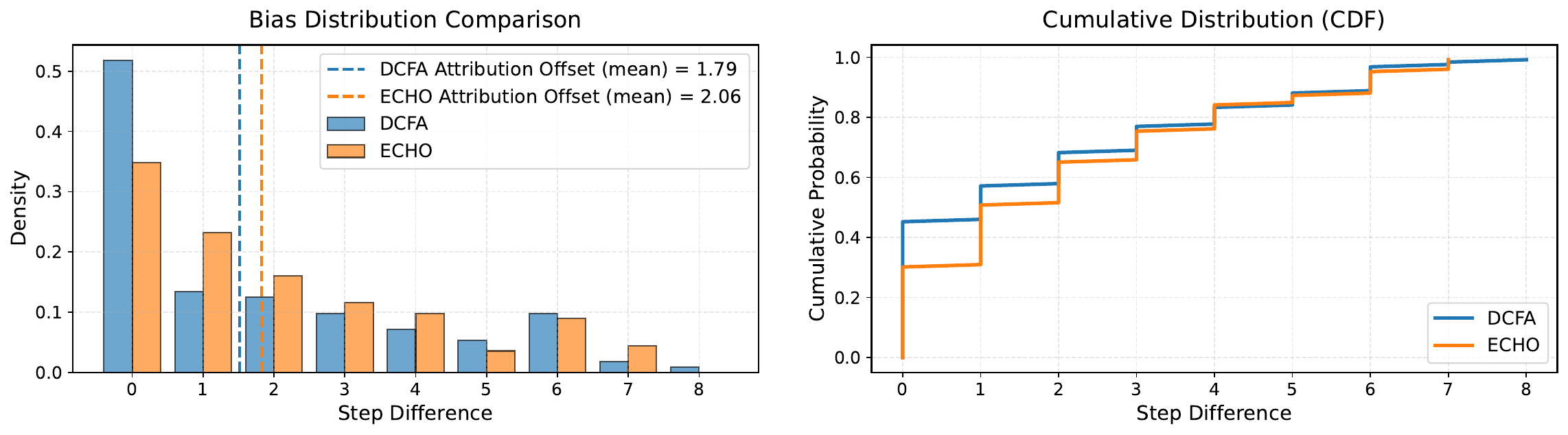}
    }
    \caption{Distribution of step-level deviations between predicted and true decisive errors. 
    DCFA shows more concentrated distributions around zero, indicating substantially reduced bias compared to ECHO.}
    \label{fig:deepseek}
\end{figure*}

\section{Bias Reduction}
\label{sec:bias_reduction}

To further validate the robustness of our causal-inspired attribution, we examine DCFA’s capability to mitigate bias in decisive error localization compared with baselines. 
Here, \emph{bias} refers to the systematic deviation of predicted steps from the ground-truth decisive error, which can obscure the actual failure source and mislead subsequent debugging~\cite{li2022training,shrotriya2025navigating}.

We compare DCFA with the representative baseline, ECHO, on both Algorithm-generated and Hand-crafted traces from the Who\&When benchmark~\cite{zhang2025agent}. 
For each trace, we quantify the deviation between the predicted and true causal-step indices:
\begin{equation}
d_{\text{DCFA}}\!=\!\left|t^{*}_{\text{DCFA}}\!-\!t^{\dag}\right|, \quad
d_{\text{Baseline}}\! =\!\left|t^{*}_{\text{Baseline}}\!-\!t^{\dag}\right|
\end{equation}
where $t^{}_{\text{DCFA}}$ and $t^{}_{\text{Baseline}}$ are predicted indices, and $t^{\dag}$ denotes the ground-truth step. A smaller deviation implies lower bias and more reliable reasoning.

Figure~\ref{fig:deepseek} visualizes the deviation distributions for DCFA and ECHO.
Across both DeepSeek-R1-671B and Gemini-2.5-pro backends, DCFA exhibits a notably sharper and more centralized distribution around zero. Overall, DCFA exhibits a smaller failure attribution offset than ECHO, indicating more accurate and less biased localization of the true decisive errors.
This indicates that DCFA predictions consistently align closer to the ground-truth decisive error.  
Such improvement stems from the both global and local view causal-inspired enhancement of the DCFA, preventing error accumulation in step-level reasoning.
Overall, the results demonstrate that DCFA reduces attribution bias, leading to more stable and interpretable failure diagnostics across synthetic and real-world traces.

\section{Related Works}
\label{app:related_works}

\subsection{LLM-based Multi-Agent Systems and System Failure}

The recent progress of Large Language Model (LLM)-based agents has accelerated the development of Multi-Agent Systems (MAS), which have emerged as a promising paradigm for addressing complex, multi-step tasks that exceed the capabilities and token limitations of single, monolithic LLMs~\cite{yang2025agentnet,sun2025llm,pei2025flow,xia2025demystifying}. 
The principles underlying MAS, such as task decomposition, inter-agent communication, and emergent collective reasoning. 
Recent surveys have proposed systematic taxonomies of agent systems architectures~\cite{AG2_2024,fourney2024magentic}, identifying core functional components required for effective coordination and execution. 
These components generally include explicit role assignment, hierarchical or iterative planning mechanisms, structured communication protocols, externalized memory modules, and evaluation and feedback mechanisms~\cite{yan2025beyond}. 

Despite rapid progress, current agent systems remain limited in robustness, interpretability, and stability~\cite{zhang2025agent,cemri2025multi}. 
Agents often exhibit brittle coordination, unstable reasoning trajectories, and cascading errors, particularly in long-horizon or highly interactive environments~\cite{zhang2025agent,cemri2025multi}. 
Benchmarking efforts such as AgentBench~\cite{liu2023agentbench} empirically demonstrate these weaknesses, showing that even state-of-the-art systems struggle to maintain coherence and consistency in collaborative, multi-step tasks. 
The recurrence of these failures exposes a critical research gap: understanding why and how LLM-based agent systems deviate from their intended execution pathways.

\subsection{Failure Attribution in LLM-based Multi-agent Systems}

{
Prior work has studied error analysis in agentic systems. AgentErrorBench with AgentDebug~\cite{zhu2025llm} locates mistakes that trigger cascading reasoning distortions in human-agent interactions in single-agent settings, whereas TRAIL~\cite{deshpande2025trail} analyzes errors in execution trajectories in multi-agent settings. These definitions, however, do not directly target the most actionable points for repairing task failures. Meanwhile, unlike single-agent systems with sequential execution and full information access, multi-agent systems involve heterogeneous roles, complex interaction topologies, partial observability, and non-trivial communication, producing intricate and interdependent dependency chains~\cite{cemri2025multi,fourney2024magentic,AG2_2024}. Thus, the focus shifts to identifying the earliest mistake whose correction could reverse overall system failure in multi-agent systems. The mistake, termed the decisive error~\cite{cemri2025multi,fourney2024magentic,AG2_2024}, represents the key point for restoring task success.
}

Recent research has begun to explore failure attribution in multi-agent systems using LLMs as diagnostic detectors~\cite{zhang2025agent,zhang2025agentracer}. 
Existing methods can be categorized into fine-tuning-based~\cite{zhang2025agentracer} and instruction-based~\cite{zhang2025agent,cemri2025multi} paradigms. 
The fine-tuning-based approach, exemplified by AgenTracer~\cite{zhang2025agentracer}, constructs specialized datasets to train dedicated models for failure diagnosis. 
While this approach offers strong controllability, it also incurs substantial costs for annotation, training, and maintenance. 
Instruction-based methods~\cite{zhang2025agent,cemri2025multi}, in contrast, rely on carefully designed prompts to guide LLMs in analyzing multi-agent system traces and identifying the decisive errors. For example, the Step-by-Step algorithm in the Who\&When benchmark~\cite{zhang2025agent} formulates failure attribution as a sequential inspection process that progressively checks each interaction.
Another algorithm in the benchmark, Binary-Search, adopts a divide-and-conquer strategy, recursively narrowing the trace scope until the decisive error is pinpointed.
Additionally, Banerjee et al.~\cite{banerjee2025did} propose hierarchical context representations and multi-perspective consensus to improve attribution reliability, while Weng et al.~\cite{west2025abduct} introduce A2P, a causal-guided attribution framework powered by the DeepScientist AI system.

Despite these advances, current methods~\cite{west2025abduct,banerjee2025did} rely primarily on the vanilla reasoning capabilities of LLMs, which often restricts their focus to superficial deviations and leads to performance degradation when processing long system traces. These limitations frequently result in inaccurate or incomplete attribution. To address these challenges, we propose a unified framework that incorporates structured reasoning and extracts local dependency chains to enable robust and reliable failure attribution in LLM-based multi-agent systems.

\subsection{Causal Reasoning and Counterfactual Analysis via LLM}

Recent advances have increasingly explored the use of large language models (LLMs) for causal reasoning and counterfactual-inspired analysis, including discovering, representing, and simulating causal relationships~\cite{su2025enhancing,cheng2025survey,luo2024open}. This line of research leverages the linguistic and reasoning capabilities of LLMs to extract and reason about causal structures from unstructured text. These studies show that LLMs can effectively extract events and propose candidate causal links with high recall, serving as powerful front-end modules for event causality analysis~\cite{liu2024identifying,wang2025event}.

To improve reliability, several works~\cite{tong2024automating,cohrs2025large,ban2025integrating,liu2025large} integrate LLM-derived evidence with algorithmic estimators, aggregating multiple causal hypotheses or graph structures and taking their intersection to enhance robustness and filter out spurious relations. This hybrid approach demonstrates that LLMs can serve as high-level causal reasoning modules, while traditional estimators provide quantitative validation. Further research explores the use of LLMs as informative priors for causal graph discovery. Recent studies demonstrate that LLMs encode rich domain and commonsense knowledge that can guide structure learning and improve interpretability~\cite{darvariu2024large,jiralerspong2024efficient}. By injecting LLM-derived constraints or edge probabilities into graph search algorithms, these methods reduce sample complexity and produce more plausible causal graphs for downstream causal analysis.

Beyond causal structure discovery, causal inference and counterfactual simulation have been explored as mechanisms for testing causal hypotheses~\cite{liu2025large,zhu2024causal,verma2025causal}. Causal inference provides formal tools for reasoning about the effects of interventions~\cite{imai2024causal}, while counterfactual simulation assesses how altering an event or reasoning step might change subsequent outcomes. Recent works apply such techniques to LLM reasoning chains, using generated interventions to evaluate faithfulness and sensitivity to hypothetical changes in chain-of-thought reasoning~\cite{tutek2025measuring,yu2024causaleval}. These findings suggest that LLMs can both represent causal structures and support the simulation of hypothetical interventions for evaluating outcome changes.

Inspired by these developments, our work adapts these ideas to failure attribution in LLM-based multi-agent systems. Rather than claiming formal causal inference, DCFA employs causal-inspired structured reasoning over system traces. Specifically, we construct a causal-inspired dependency graph to organize inter-agent dependencies and use counterfactual-inspired evaluation with LLM-mediated approximate interventions to assess how correcting candidate interactions may alter the final outcome. This dual-view design provides an interpretable basis for identifying the interaction that most plausibly constitutes the decisive error in long and unstructured multi-agent traces.

\begin{algorithm}[!b]
\caption{Global Causal-inspired Attribution (GCA)}
\label{alg:gca}
{
\begin{algorithmic}
\STATE {\bfseries Input:} trace $\mathcal{T}$, ground truth $\widehat{\mathcal{Y}}$
\STATE {\bfseries Output:} hypothesis $(\tau^{\star}, r^{\star})$, causal-inspired dependency graph $\mathcal{G}$

\STATE Minor Deviation Identification
\STATE $(\mathcal{T}^{e},\mathcal{R}^{e}) \leftarrow \mathrm{MDI}(\mathcal{T})$

\STATE Dependency Graph Construction
\STATE $\mathcal{G} \leftarrow 
\mathrm{CGC}(\mathcal{T},\mathcal{T}^{e},\mathcal{R}^{e})$

\STATE Hypothesis Generation
\STATE $(\tau^{\star}, r^{\star}) \leftarrow 
\mathrm{HG}(\widehat{\mathcal{Y}},\mathcal{T},\mathcal{T}^{e},\mathcal{R}^{e},\mathcal{G})$

\STATE {\bfseries return} $\tau^{\star}, r^{\star},\mathcal{G}$
\end{algorithmic}
}
\end{algorithm}

\section{Pseudocode Description of DCFA}
\label{sec:appendix_algorithm}

This section briefly explains the pseudocode of the two modules in the DCFA framework.

\paragraph{Global Causal-inspired Attribution (GCA).}
Algorithm~\ref{alg:gca} summarizes the overall global attribution procedure. The module first identifies candidate minor deviations in the system trace using the deviation identification process $\mathrm{MDI}(\cdot)$, which extracts interactions that exhibit abnormal behaviors together with brief textual explanations. Based on these candidates, the causal-inspired dependency graph construction procedure $\mathrm{CGC}(\cdot)$ organizes the trace into a directed causal-inspired dependency graph that encodes plausible dependencies between interactions. Finally, the hypothesis generation process $\mathrm{HG}(\cdot)$ analyzes the trace, the detected deviations, and the constructed causal-inspired dependency graph to produce an initial hypothesis $(\tau^{\star}, r^{\star})$ for the decisive error.

\begin{algorithm}[!b]
\caption{Local Counterfactual-inspired Enhancement (LCE)}
\label{alg:lce}
{
\begin{algorithmic}
\STATE {\bfseries Input:} trace $\mathcal{T}$, ground truth $\widehat{Y}$, causal-inspired dependency graph $\mathcal{G}$, hypothesis $\tau^{\star}$
\STATE {\bfseries Output:} decisive error $\tau^{*}$

\STATE $\widehat{\mathcal{T}} \leftarrow \mathrm{Correct}(\widehat{Y},\mathcal{T},\mathcal{G})$
\STATE $p \leftarrow \text{index}(\tau^{\star})$

\STATE Forward Search
\WHILE{True}
    \STATE $\mathcal{N} \leftarrow \{\tau_j \mid (p,j)\in\mathcal{E}\}$
    \STATE evaluate $\Delta\mathcal{S}(\tau_j)$ for $\tau_j\in\mathcal{N}$
    \STATE $p' \leftarrow \arg\max_{\tau_j\in\mathcal{N}}\Delta\mathcal{S}(\tau_j)$
    \IF{$\Delta\mathcal{S}(p')>0$}
    \STATE $p \leftarrow p'$
\ELSE
    \STATE \textbf{break}
\ENDIF
\ENDWHILE

\STATE $p \leftarrow \text{index}(\tau^{\star})$

\STATE Backward Search
\WHILE{True}
    \STATE $\mathcal{N} \leftarrow \{\tau_k \mid (k,p)\in\mathcal{E}\}$
    \STATE evaluate $\Delta\mathcal{S}(\tau_k)$ for $\tau_k\in\mathcal{N}$
    \STATE $p' \leftarrow \arg\max_{\tau_k\in\mathcal{N}}\Delta\mathcal{S}(\tau_k)$
    \IF{$\Delta\mathcal{S}(p')>0$}
    \STATE $p \leftarrow p'$
\ELSE
    \STATE \textbf{break}
\ENDIF
\ENDWHILE

\STATE $\mathcal{T}^{c} \leftarrow$ visited interactions
\STATE $\tau^{*} \leftarrow \mathrm{DED}(\widehat{Y},\mathcal{T}^{c},\mathcal{G})$

\STATE {\bfseries return} $\tau^{*}$
\end{algorithmic}
}
\end{algorithm}

\paragraph{Local Counterfactual-inspired Enhancement (LCE).}
Algorithm~\ref{alg:lce} refines the global hypothesis by performing a bidirectional greedy search on the causal-inspired dependency graph. The procedure first constructs a corrected trace using $\mathrm{Correct}(\cdot)$ to provide a reference trajectory aligned with the ground-truth outcome. Starting from the hypothesis interaction $\tau^{\star}$, the algorithm iteratively explores downstream and upstream neighbors in two search loops. At each step, candidate interactions are evaluated using counterfactual-inspired interventions, and the interaction with the largest positive marginal improvement $\Delta\mathcal{S}$ becomes the next pivot. The union of visited interactions forms a compact dependency neighborhood, which is then evaluated by the decision function $\mathrm{DED}(\cdot)$ to determine the final decisive error.

\section{Prompts for DCFA}
\label{sec:appendix_prompts}

This section presents the prompts used to implement the DCFA components described in Section~\ref{sec:methodology}.
The prompts correspond to key reasoning steps in the Global Causal-inspired Attribution (GCA) and Local Counterfactual-inspired Enhancement (LCE) modules.
For clarity and reproducibility, we provide representative prompt templates. Actual inputs (e.g., system traces, candidate events, and ground-truth answers) are dynamically inserted during execution.

\paragraph{Prompts for the GCA Module.}

The GCA module performs global structured reasoning over the full system trace to generate an initial hypothesis for the decisive error.

\begin{figure*}[t]
    \centering
    \includegraphics[width=0.95\textwidth]{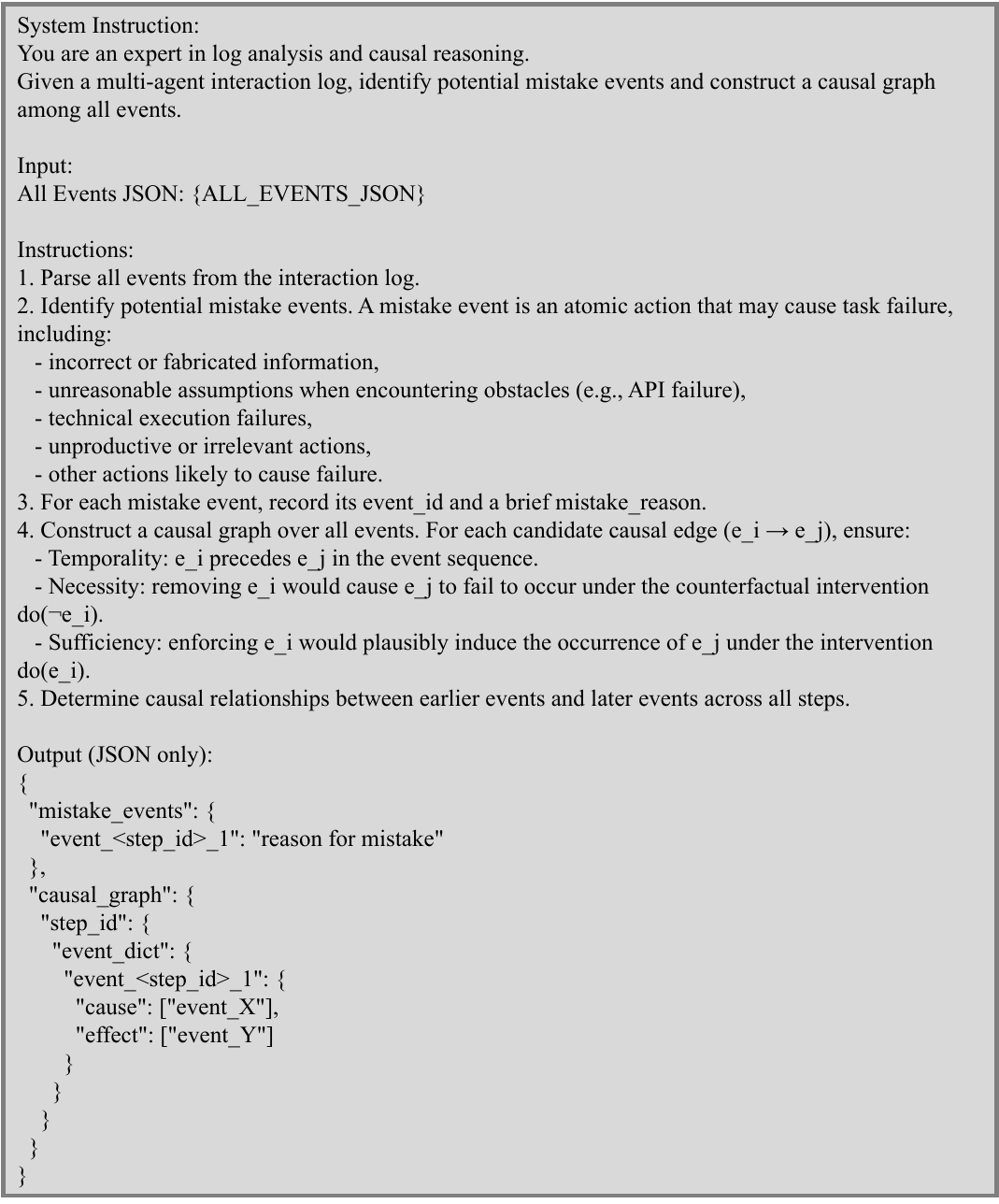}
    \caption{
    Prompt used for minor deviation identification and causal-inspired dependency graph construction. 
    The model identifies surface-level mistake events as minor deviations, 
    and establishes dependency relations among them according to temporality, necessity, and sufficiency criteria.
    }
    \label{fig:prompt_gca1}
\end{figure*}

First, the model identifies candidate \textit{minor deviations} and constructs a causal-inspired dependency graph over the system trace.
The prompt instructs the model to analyze the interaction events extracted from the system trace, detect potential mistake events, and identify dependency relationships among events based on the criteria of \textit{temporality}, \textit{necessity}, and \textit{sufficiency} defined in Section~\ref{sec:GCA}.
This step jointly performs minor deviation identification and causal-inspired dependency graph construction, producing both the set of candidate deviation events and the directed dependency relationships among them.
The prompt used for this step is shown in Figure~\ref{fig:prompt_gca1}.

\begin{figure*}[!h]
    \centering
    \includegraphics[width=0.95\textwidth]{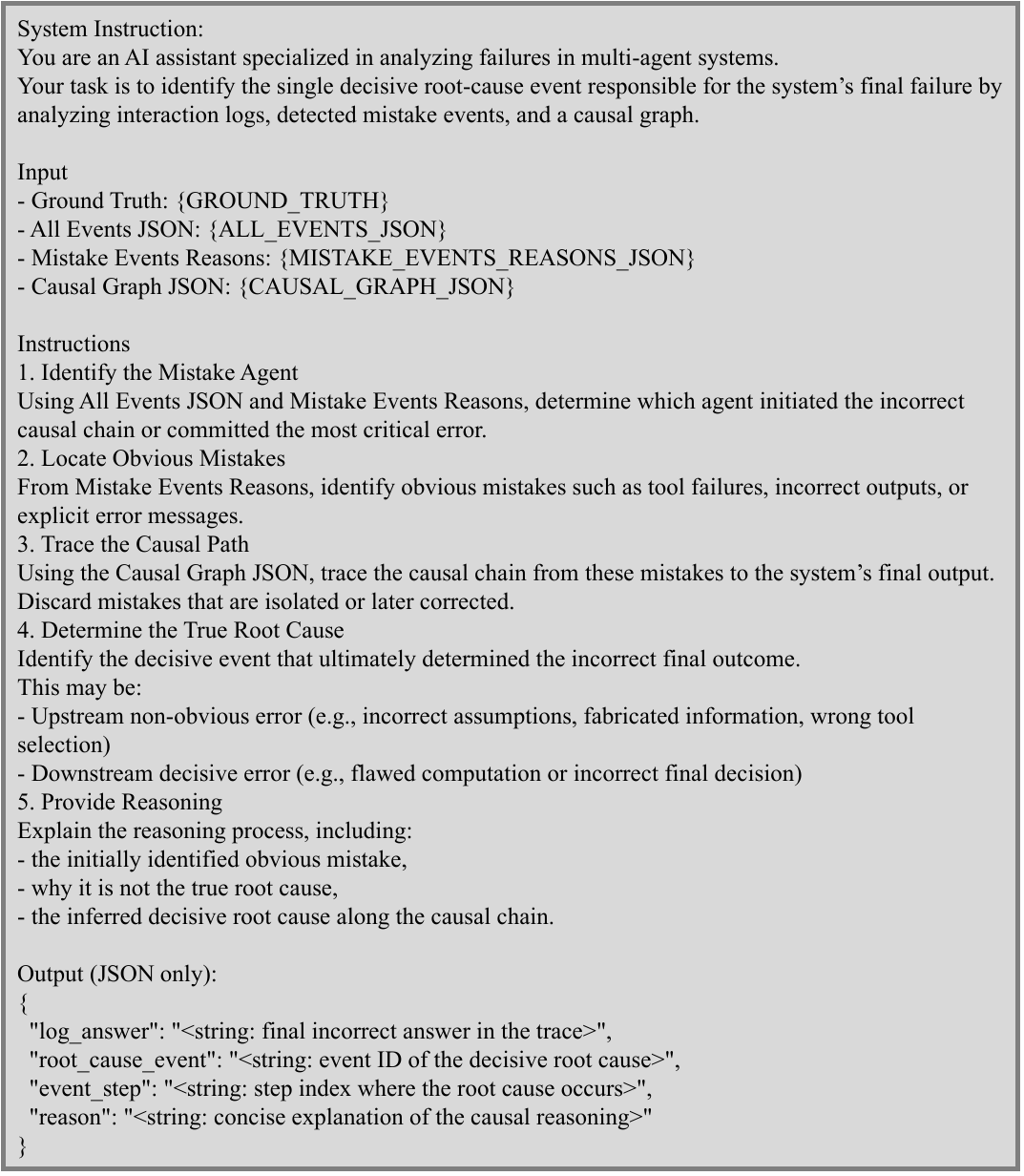}
    \caption{
    Prompt used for global causal-inspired structured reasoning and hypothesis generation. Given the system trace, candidate minor deviations, their explanations, and the constructed causal-inspired dependency graph, the model performs global reasoning over the dependency structure to analyze how errors may propagate and proposes an initial hypothesis for the decisive error.
    }
    \label{fig:prompt_gca2}
\end{figure*}

After the candidate deviations and causal-inspired dependency graph are obtained, the GCA module performs global reasoning to generate an initial hypothesis of the decisive error. 
Given the system trace, the set of candidate minor deviations, the explanations for each deviation, and the constructed causal-inspired dependency graph, the model evaluates how deviations propagate through the interaction chain and selects the event that most plausibly explains the final failure outcome. 
The corresponding prompt is illustrated in Figure~\ref{fig:prompt_gca2}.

\paragraph{Prompts for the LCE Module.}

The LCE refines the hypothesis generated by GCA through localized counterfactual-inspired reasoning, comprising prompt-based steps for error selection.

\begin{figure*}[!h]
    \centering
    \includegraphics[width=0.95\textwidth]{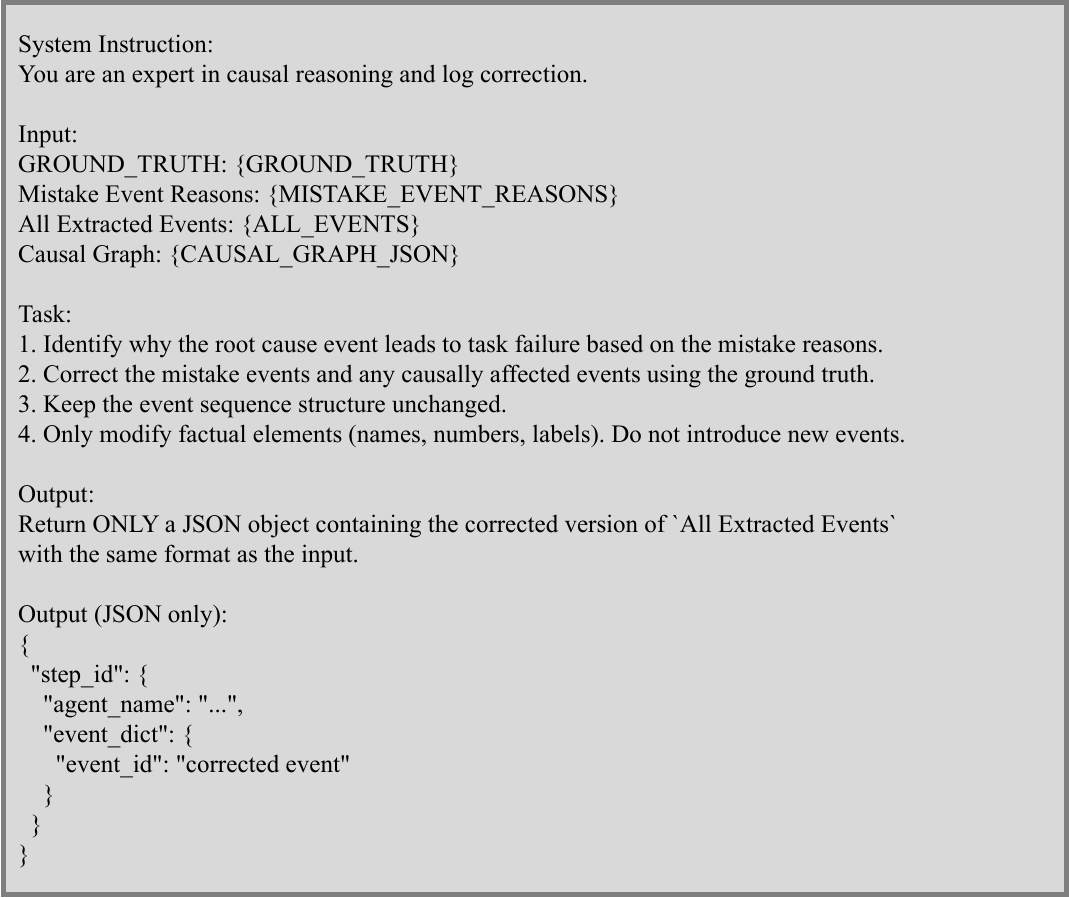}
    \caption{
    Prompt used for causal-inspired log correction. 
    The model corrects mistake events and causally affected events using ground-truth 
    information while preserving the original event structure and format.
    }
    \label{fig:prompt_lce1}
\end{figure*}

First, in the counterfactual-inspired evaluation process, the model generates a corrected system trace.
Given the original trace, the identified mistake events, and the ground-truth outcome, the prompt instructs the model to correct the erroneous events and causally affected downstream interactions while preserving the original event structure and formatting.
This produces a fully corrected trace representing a counterfactual-inspired scenario in which the mistakes are fixed.
The prompt is shown in Figure~\ref{fig:prompt_lce1}.

\begin{figure*}[!h]
    \centering
    \includegraphics[width=0.95\textwidth]{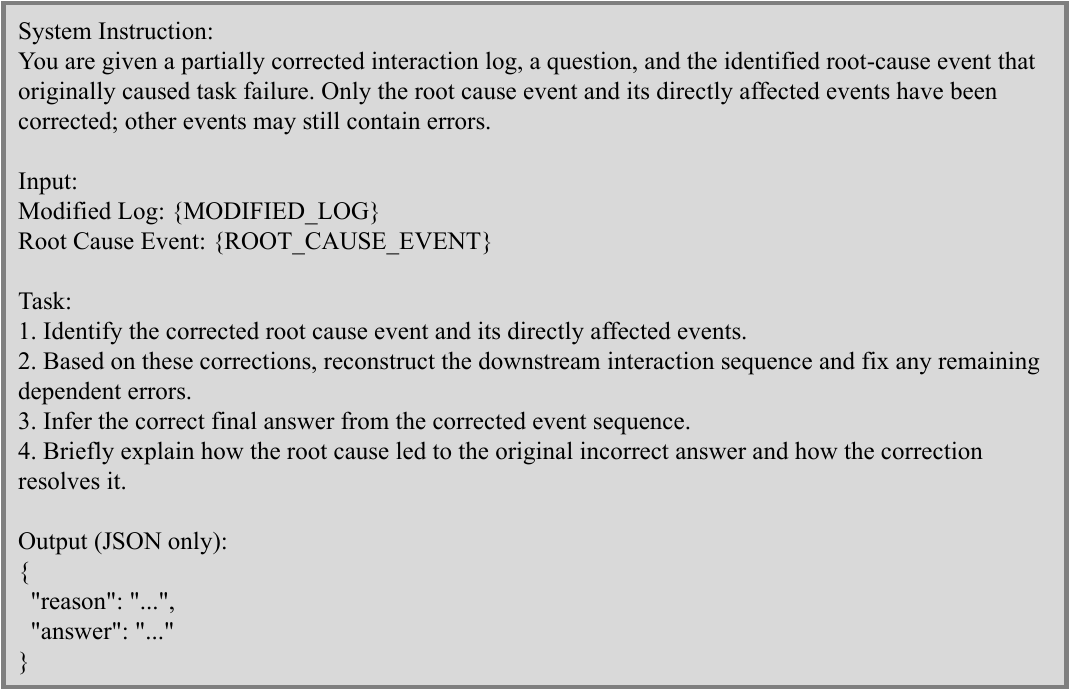}
    \caption{
    Prompt used for counterfactual-inspired outcome simulation. 
    Given a partially corrected interaction trace, the model reconstructs the downstream reasoning process 
    and predicts the resulting final outcome.
    }
    \label{fig:prompt_lce2}
\end{figure*}

Next, for each counterfactual-inspired trace constructed during evaluation, the model simulates the corresponding system outcome based on the partially corrected interaction sequence.
Starting from the corrected prefix and the remaining original interactions, the model reconstructs the downstream reasoning process and predicts the resulting final outcome.
This prompt enables the model to estimate how correcting specific interactions affects the final result.
The corresponding prompt is presented in Figure~\ref{fig:prompt_lce2}.

\begin{figure*}[!h]
    \centering
    \includegraphics[width=0.95\textwidth]{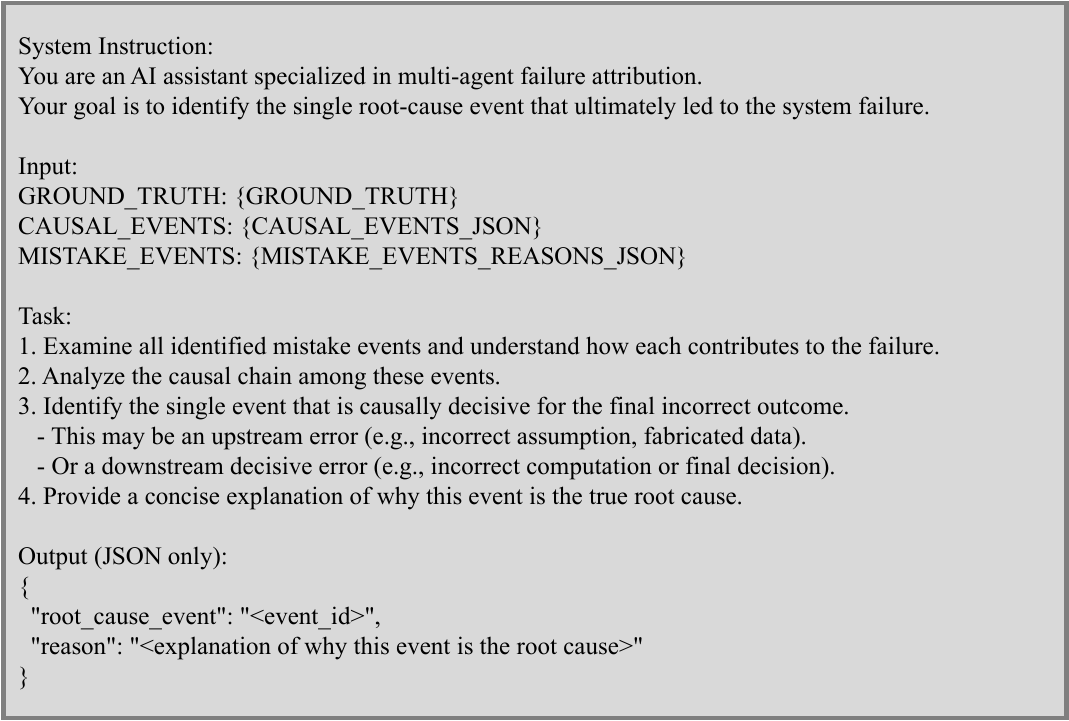}
    \caption{
    Prompt used for decisive error identification. 
    The model analyzes candidate mistake events and selects the single event that 
    is causally decisive for the final system failure.
    }
    \label{fig:prompt_lce3}
\end{figure*}

Finally, after the bidirectional dependency neighborhood search and counterfactual-inspired evaluation produce a set of candidate decisive error events, the model performs a final semantic decision and selection step.
Given the candidate events, the causal-inspired dependency graph structure, and the ground-truth outcome, the model determines which interaction most plausibly constitutes the decisive error responsible for the observed system failure.
The prompt is shown in Figure~\ref{fig:prompt_lce3}.

\end{document}